\documentclass[runningheads]{llncs}

\usepackage{eccvabbrv}

\usepackage{graphicx}
\usepackage{booktabs}
\usepackage{enumitem}
\usepackage{wrapfig}
\usepackage{cite}
\usepackage{subcaption}
\usepackage{float}
\usepackage[accsupp]{axessibility}  

\usepackage{hyperref}
\usepackage{multirow}
\usepackage{orcidlink}
\usepackage[most]{tcolorbox}
\definecolor{darkgreen}{RGB}{0,100,0} 
\newcommand{\methodname}{KineMask\xspace}

\newcommand{\interactions}{\textit{Interactions}\xspace}

\begin{document}

\title{Learning to Generate Rigid Body Interactions with Video Diffusion Models} 

\titlerunning{Learning to Generate Rigid Body Interactions with VDMs}

\author{David Romero\inst{1} \and
Ariana Bermudez\inst{1} \and
Viacheslav Iablochnikov\inst{1}\\
Hao Li\inst{1,2} \and
Fabio Pizzati\inst{1} \and
Ivan Laptev\inst{1}}

\authorrunning{D. Romero et al.}

\institute{MBZUAI, UAE \and
Pinscreen, USA \\
\email{\{david.romero, fabio.pizzati,  ivan.laptev\}@mbzuai.ac.ae}}
\maketitle

\begin{abstract}
  Recent video generation models have achieved remarkable progress and are now deployed in film, social media production, and advertising. Beyond their creative potential, such models also hold promise as world simulators for robotics and embodied decision making. Despite strong advances, current approaches still struggle to generate physically plausible object interactions and lack object-level control mechanisms. To address these limitations, we introduce \emph{KineMask}, an approach for video generation that enables realistic rigid body control, interactions, and effects. Given a single image and a specified object velocity, our method generates videos with inferred motions and future object interactions. We propose a two-stage training strategy that gradually removes future motion supervision via object masks. Using this strategy we train video diffusion models (VDMs) on synthetic scenes of simple interactions and demonstrate significant improvements and generalization to rigid body and hand-object interactions in real scenes. Furthermore, \emph{KineMask} integrates low-level motion control with high-level textual conditioning via predicted scene descriptions, leading to support for synthesis of complex dynamical phenomena.
Our experiments show that \emph{KineMask} generalizes to different VDMs and achieves strong improvements over recent models of comparable size. Ablation studies further highlight the complementary roles of low- and high-level conditioning in VDMs. 

Project Page: \url{https://daromog.github.io/KineMask/}
  \keywords{Object Interactions \and Motion Conditioning Generation}
\end{abstract}

\section{Introduction}
\label{sec:intro}

Recent years have seen substantial advances in video generation, with Video Diffusion Models (VDMs) emerging as a leading paradigm for high-resolution, temporally consistent synthesis~\cite{blattmann2023align, ho2020denoising, kong2024hunyuanvideo, yang2024cogvideox}. This progress has elevated visual quality and enabled early commercial use in creative performances~\cite{miller2023elvis}, experimental filmmaking \cite{newyorker_trillo2023}, and advertising~\cite{theverge_kalshi_ai_ad_2025}. Beyond content creation, VDMs have also been explored as world models ~\cite{ha2018world}, capable of anticipating real-world interactions and supporting robotics and embodied decision making~\cite{agarwal2025cosmos,alonso2024diffusion,ding2024diffusion}. Realizing this vision, however, requires strict physical plausibility and control, since small deviations from realistic physics can accumulate into large errors in predicted dynamics. Yet, current VDMs struggle to capture fundamental traits~\cite{motamed2025generative,kang2024far} such as object permanence, collisions and causal interactions, even for large-scale models like Veo-3~\cite{veostudy}. Hence, methods for physically accurate video generation are a key step toward establishing VDMs as reliable world models.
\begin{figure}[t]
  \newcommand{\teaserwidth}{\textwidth}
  \centerline{
    \includegraphics[width=\teaserwidth]{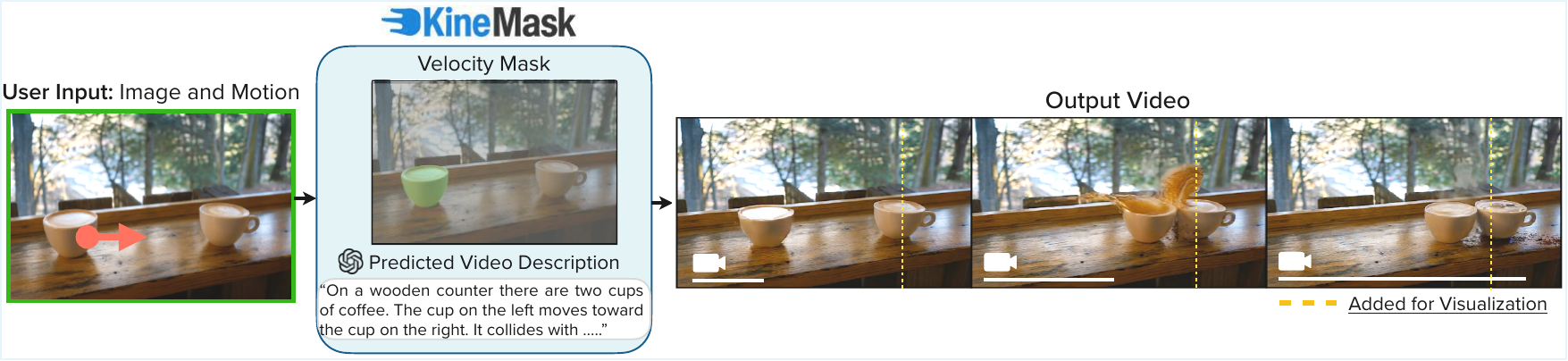}
    }
    \captionof{figure}{We enable object-based control with a novel training strategy. Paired with synthetic data constructed for the task, \textbf{\emph{KineMask}} enables pretrained VDMs to synthesize realistic rigid body interactions in real-world input scenes.}
    \vspace{-14pt}
  \label{fig:teaser}
 \end{figure}
Recent frameworks such as WonderPlay~\cite{li2025wonderplay}, PhysGen ~\cite{liu2024physgen} and PhysGen3D~\cite{chen2025physgen3d}  integrate physics-based simulators into data-driven video generation, but rely on explicit scene reconstruction, a challenging task on its own. Differently, drag-based approaches~\cite{shi2023dragdiffusion,li2023motionctrl,zhang2025tora,geng2025motionprompt,zhou2025trackgo,lei2025animateanything,chen2023motion} can move objects, but often requiring pre-defined target points, thus preventing the inference of causal effects of motion from starting conditions only. Around the time of this work, Force Prompting~\cite{gillman2025force} introduced physics-guided controls into VDMs using simulated training data. While promising, it offers limited fine-grained control and object interactions realism, often producing unrealistic dynamics, and distorted shapes, showing a lack of understanding of causality. Motivated by these limitations, we propose an approach enabling video diffusion models to generate physically realistic object interactions, as a core requirement for robotics and video generation. Our approach aims to answer two central questions for world models: \textbf{(1)} Can a video diffusion model generate \textit{realistic interactions} between rigid bodies given initial dynamic conditions?, and \textbf{(2)} how do data and textual conditioning influence the emergence of causal physical effects in generated videos?.\looseness=-1\\ 
\indent To address these questions, we introduce \emph{KineMask}, a framework for generating accurate object interactions and effects in complex scenes (Figure~\ref{fig:teaser}). \emph{KineMask} provides \textit{low-level} kinematic control over parameters such as object direction and speed, while \textit{inferring object interactions directly within the VDM}. \emph{KineMask} is trained on simulator-rendered videos that capture not only physically valid dynamics but also explicit speed-dependent object interactions, paired with textual descriptions of the underlying events. We leverage object masks as guidance, enabling single-frame motion control and improving the model’s understanding of shapes. Beyond low-level control, \emph{KineMask} integrates \textit{high-level} prompt conditioning through textual descriptions of future scene dynamics. KineMask provides low-level Kinematic control through velocity masks, combined with high-level text conditioning, and learns physically-inspired causality and motion from simulated data, improving controllability and physical plausibility. At inference, it predicts dynamics from a single input image and enables the generation of complex effects, such as liquid spilling as a result of a collision (Figure~\ref{fig:teaser}). Extensive experiments show that \emph{KineMask} not only introduces new control capabilities but also outperforms state-of-the-art models of comparable size, while ablation studies highlight the importance of both the proposed training strategy and the integration of low- and high-level controls. In addition, we show that improvements of \emph{KineMask} generalize to different VDMs. To this end, we first demonstrate advantages of \emph{KineMask} for \emph{CogVideoX-5B}~\cite{yang2024cogvideox} and then verify improvements by applying \emph{KineMask} to \emph{Wan2.2-5B}~\cite{wan2025wan} and \emph{Cosmos2.5-2B}~\cite{agarwal2025cosmos} video models. In summary, we propose the following contributions:\looseness=-1

\begin{enumerate}
    \item We introduce KineMask, a mechanism for object motion conditioning in VDMs, based on a novel two-stage training and conditioning encoding.
    \item We train KineMask on a synthetic video dataset composed of simple object interactions and demonstrate resulting models to enable generation of complex object interactions in realistic scenes.
    \item We evaluate KineMask by combining low-level motion control with high-level textual conditioning, and demonstrate significant improvements of multiple video models in the task of generating realistic rigid body interactions.
\end{enumerate}

\section{Related Works}

\vspace{-3px}\subsection{Video Diffusion Models}
Early VDMs directly extended image generators by inserting temporal layers into denoising U-Nets~\cite{ho2022video,blattmann2023stable,guo2023animatediff,wang2023modelscope,wang2023lavie,chen2023videocrafter,bar-tal2024lumiere}. Later, video synthesis improved through the usage of Diffusion Transformers (DiTs)~\cite{peebles2023dit}, inheriting scaling properties from native transformer-based architectures. The use of DiTs allowed for higher-resolution videos and stronger visual quality~\cite{yang2024cogvideox,kong2024hunyuanvideo,wan2025wan, hacohen2024ltx}. We build \emph{KineMask} on CogVideoX~\cite{yang2024cogvideox}, benefiting from the advantages of DiTs. Beyond creative purposes, diffusion-based video generation is now used for \emph{world modeling}, by synthesizing the possible outcomes of actions in an environment. \cite{alonso2024diffusion,valevski2024diffusion} first proposed diffusion-based world models on restricted scenarios. Some like GAIA~\cite{hu2023gaia,russell2025gaia} train at scale on specific domains such as autonomous driving, while Cosmos~\cite{agarwal2025cosmos} uses large-scale data to handle heterogeneous domains. These models still suffer from limited realism in generating object interactions, motivating our study.\looseness=-1

\vspace{-5pt}\subsection{Control for video generation}
Significant efforts have been made to extend control on generated videos beyond textual conditioning. 
On images, ControlNet~\cite{zhang2023adding}  and similar approaches~\cite{mou2023t2iadapter,zhao2023unicontrol} proposed plug-and-play control trainable components for dense conditioning such as edges, depth, or human poses.
In videos, VideoControlNet~\cite{hu2023videocontrolnet} propagated conditions frame by frame using optical flow, while others focused on pose-driven human video generation~\cite{zhang2024mimicmotion} or keyframe-to-video propagation~\cite{li2024keyvid}. 
More recent works such as SparseCtrl~\cite{guo2023sparsectrl} uses just a few keyframes for sketch, depth, or image conditioning, ignoring motion conditions. MotionI2V~\cite{shi2024motion} first infers object motion with flow generation, then it renders a video using flow as control.
A significant line of works allows motion control with dragging trajectories~\cite{shi2023dragdiffusion,li2023motionctrl,zhang2025tora,geng2025motionprompt,zhou2025trackgo,lei2025animateanything,deng2024dragvideo, wan2024dragentitytrajectoryguidedvideo,yin2023dragnuwa}. While effective, these require pre-defined trajectories, and are unable to infer motion causal effects from initial conditions. Some drive motion by directly guiding attentions~\cite{pondaven2025video} or noise~\cite{burgert2025go} using reference videos. Beyond single modalities, Cosmos-Transfer~\cite{abu2025cosmostransfer} demonstrates adaptive multimodal conditioning. To our knowledge, \emph{KineMask} is the first approach training VDMs to generate rigid body interactions by controlling \textit{initial object velocity only}.\looseness=-1

\begin{figure*}[t]
    \centering
    \includegraphics[width=\linewidth]{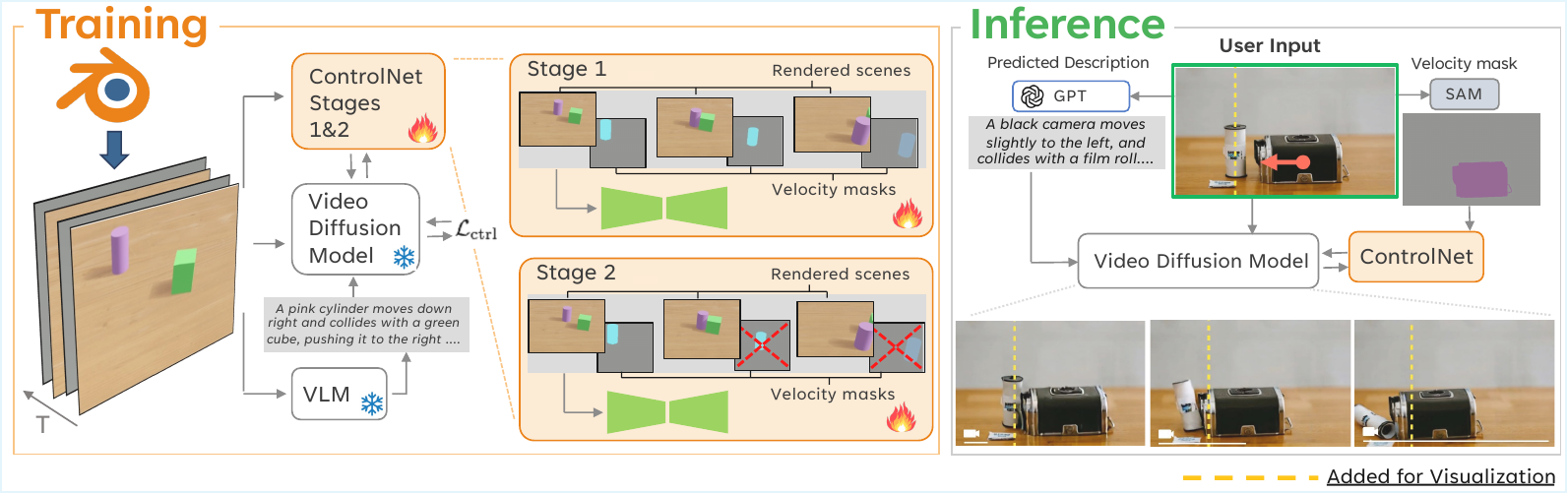}
    \caption{\textbf{KineMask pipeline.} We encode our \textit{low-level} control signal as a mask encoding the velocity of the moving objects, to train a ControlNet (left) in two stages using Blender-generated videos of objects in motion. In the first stage, we train with all mask frames as control. In the second stage, we randomly drop parts of the final mask frames. Additionally, we provide a \textit{high-level} textual control extracted by a VLM. At inference (right), we construct the low-level conditioning with SAM and use GPT to infer high-level outcomes of object motion from a single frame.}\vspace{-10pt}
    \label{fig:Architecure}
\end{figure*}

\vspace{-5pt}\subsection{Physics-aware video generation}
The interactions between physical understanding and video generation is a growing research field. A first line of work integrates physical simulations with learning-based techniques. In particular, PhysDreamer~\cite{zhang2024physdreamer} generates oscillatory motions on 3D Gaussians, while
DreamPhysics learns physical properties of dynamic 3D Gaussians with video diffusion priors~\cite{huang2025dreamphysics}. On video generation, WonderPlay~\cite{li2025wonderplay} bridges physics solvers and generative video models to synthesize dynamic 3D scenes across diverse physical phenomena. PhysGen3D~\cite{chen2025physgen3d} and PhysGen~\cite{liu2024physgen} apply a similar approach using 3D and 2D information respectively for rigid body interactions. However, the use of a simulator requires significant engineering efforts and limits the flexibility of models. Differently, others~\cite{blattmann2021ipoke,davtyan2024learn} map input kinematics to future frames, but requiring dataset-specific training. Instead, C-Drag~\cite{li2025c} infers causal motion on output videos with LLMs, and combines it with tracking control to allow video generation. Using purely diffusion models, InterDyn~\cite{akkerman2025interdyn} explores the capability of video models to render realistic object dynamics. However, it exploits frame-wise masks of controlling elements, which are typically unavailable at test time. Similarly to us, \cite{li2025pisa} explores physics post-training but focuses on gravity effects. Goal Force \cite{gillman2026goalforceteachingvideo} focuses on goal forces and Force Prompting~\cite{gillman2025force} explores similar ideas to ours but does not consider object interactions and deploys simpler motion control. We compare to Force Prompting and demonstrate improved performance synthesizing object interactions.\looseness=-1

\vspace{-8px}\section{KineMask}
KineMask is designed to synthesize realistic interactions among rigid bodies given an input image and the object velocity encoded by an object mask, as in Figure~\ref{fig:Architecure}. Below, we introduce preliminaries in Section~\ref{sec:preliminaries}. We then describe our conditioning mechanism in Section~\ref{sec:motion-control} and outline the training and inference procedure in Section~\ref{sec:data}.\looseness=-1

\vspace{-5px}\subsection{Preliminaries}\label{sec:preliminaries}
\subsubsection{Video diffusion Models.} VDMs generate data by reversing the noising process. In training, a clean video $\mathbf{x}_{0} \in \mathbb{R}^{F \times H \times W \times C}$ is perturbed into a noisy version $\mathbf{x}_{t}$ by adding Gaussian noise at a randomly sampled timestep $t$, and the model is optimized to approximate the corresponding reverse transition. At inference, the process is inverted: starting from pure Gaussian noise $\mathbf{x}_{T}$, the model denoises through intermediate states $\{\mathbf{x}_{t}\}_{t=1}^{T}$ until it recovers a clean output video $\mathbf{x}_{0}$ after $T$ steps. To be grounded to real scenes, we use image-to-video (I2V) models, where the video synthesis is conditioned on a reference image $\mathbf{y}$. Formally, we denote by $p_{\theta}$ the VDM with parameters $\theta$. Conditioned on a high-level textual description $c$ and a reference image $\mathbf{y}$, the denoising step is:
\vspace{-2px}\begin{equation}
\mathbf{x}_{t-1} \sim p_{\theta}(\mathbf{x}_{t-1} \mid \mathbf{x}_{t}, c, \mathbf{y}),
\text{\vspace{-2px}}\end{equation}
where $\mathbf{x}_{t-1}$ is a tensor defined over the frame dimensions. The training loss minimizes the KL divergence between the true reverse conditional $p$ and the model distribution:
\vspace{-2px}\begin{equation}
    \begin{split}
    &\mathcal{L}_{\text{diff}}(\theta; \mathbf{x}_{0}, \mathbf{y}, c, t) 
    = D_{\mathrm{KL}}\!\big(p(\mathbf{x}_{t-1}\mid \mathbf{x}_{t}, \mathbf{x}_{0}) \;\big\|\; 
    p_{\theta}(\mathbf{x}_{t-1}\mid \mathbf{x}_{t}, c, \mathbf{y}) 
    \big).
    \text{\vspace{-2px}}
    \end{split}
\end{equation}

\noindent In practice, this objective is implemented by a noise prediction task in which the network learns to estimate the Gaussian noise added to $\mathbf{x}_{0}$.  
\vspace{-5px}\subsubsection{ControlNet.} To allow additional guidance, a ControlNet~\cite{zhang2023adding} branch $\psi_{\phi}$, parameterized by $\phi$, can encode an arbitrary dense control signal $\mathbf{u} \in \mathbb{R}^{F \times H \times W \times D}$ driving the output generation. For more details, please refer to~\cite{zhang2023adding}. The denoising step then becomes:
\begin{equation}
\mathbf{x}_{t-1} \sim p_{\theta}\!\bigl(\mathbf{x}_{t-1} \mid \mathbf{x}_{t}, c, \mathbf{y}, \psi_{\phi}(\mathbf{u})\bigr).
\end{equation}
When training the ControlNet, the parameters of the backbone model $\theta$ are kept frozen, while only the control branch $\phi$ is optimized. The corresponding loss is:
\begin{equation}
\begin{split}
&\mathcal{L}_{\text{ctrl}}(\phi;\mathbf{x}_{0},\mathbf{y},\mathbf{u},c,t)
= D_{\mathrm{KL}}\!\big(q(\mathbf{x}_{t-1}\!\mid\!\mathbf{x}_{t},\mathbf{x}_{0})
\,\big\|\,
p_{\theta}(\mathbf{x}_{t-1}\!\mid\!\mathbf{x}_{t},c,\mathbf{y},\psi_{\phi}(\mathbf{u}))
\big).
\end{split}
\end{equation}

\vspace{-10px}\subsection{Enabling Motion Control}\label{sec:motion-control}
\subsubsection{First-Stage training.}
We now want to enable object-wise motion control for KineMask. Specifically, our goal is to move an object in an input scene $\mathbf{y}$ with controlled direction and velocity, allowing us to study the effects of object interactions in the videos generated with diffusion models. To do so, we assume access to a dataset $\mathcal{D}$ of captioned videos depicting objects in motion. Let $f \in \{1,\dots,F\}$ denote the frame index. For each frame $f$ and object in the scene, we are given a mask $\mathbf{m}_{f} \in \mathbb{R}^{H \times W \times 3}$ aligned with the image resolution. The three channels encode the instantaneous velocity vector, with the red, green, and blue channels corresponding to motion along the $x$-, $y$-, and $z$-axes, respectively, in the pixels defined by a segmentation mask of the object. In this way, $\mathcal{D}$ provides not only spatial information about object locations, but also explicit ground-truth dynamics. The velocity masks are then aggregated into a tensor $\mathbf{m} \in \mathbb{R}^{F \times H \times W \times 3}$ and used to condition $\psi_{\phi}$. Similarily to~\cite{akkerman2025interdyn}, \textit{we annotate in this way only the velocity of the objects moving in the first frame of the rendered video}, leaving blank the mask for objects potentially moved by interactions. We visualize our strategy in Figure~\ref{fig:Architecure} (top). This enforces the model to synthesize interactions without explicitly relying on pixel control. We can then train a \textit{first-stage} KineMask ControlNet $\phi'$ by solving: 
\begin{equation}
\phi' = \arg\min_{\phi} \;\; \mathbb{E}_{(\mathbf{x}_{0}, \mathbf{y}, \mathbf{m}, c)\sim \mathcal{D},t} \Big[ \mathcal{L}_{\text{ctrl}}(\phi; \mathbf{x}_{0}, \mathbf{y}, \mathbf{m}, c, t) \Big],
\end{equation}
The $\phi'$ network learns to map dense pixel-wise supervision into structured guidance for object motion in the generated videos. We show the training masks in Figure~\ref{fig:Architecure} (top), more examples can be seen in Appendix.

\vspace{-10px}\subsubsection{Second-Stage Training}
KineMask $\phi'$ enables motion control given motion masks $\mathbf{m}$ provided for all the frames of a video. While such a setup simplifies training, it does not correspond to our desired scenario of video generation conditioned {\em only} by the object motion at the first video frame.
Towards this goal, we use a mask dropout strategy, randomly erasing part of the velocity masks in $\mathbf{m}$ at training time, as shown in Figure~\ref{fig:Architecure} (bottom). Formally, we define a truncated mask tensor:
\begin{equation}
\mathbf{m}_\odot =\{\mathbf{m}_{\odot, f} = \mathbf{m}_{f} \;\; \text{if } f \in f^{*}, \;\; \mathbf{0} \;\; \text{otherwise}\}_{f=0}^F.
\end{equation}
where $f^{*}$ denotes the masked frame indices, and depends on the dropout ratio. Thus, only some frames contain velocity supervision, while the remainder of the sequence is set to zero. We then train a \textit{second-stage} KineMask $\phi''$ by finetuning $\phi'$ with this strategy, solving:
\begin{equation}
\phi^{\prime\prime} = \arg\min_{\phi''} \;\; \mathbb{E}_{(\mathbf{x}_{0}, \mathbf{y}, \mathbf{m},c), \sim \mathcal{D}, t} \Big[ \mathcal{L}_{\text{ctrl}}(\phi'; \mathbf{x}_{0}, \mathbf{y}, \mathbf{m}_{\odot}, c, t) \Big].
\end{equation}
As a result of the dropout during training, the VDM equipped with $\phi''$ is able to move objects by taking as input only the \textit{initial} velocity, with $\mathbf{m}_\odot=\{\mathbf{m}_0, \mathbf{0}, ..., \mathbf{0}\}$. Ultimately, to render realistic videos, the VDM \textit{must synthesize motion dynamics} starting from initial conditions only.

\vspace{-5px}\subsection{Data}\label{sec:data}
\subsubsection{Training.}
We show our training pipeline in Figure~\ref{fig:Architecure} (left). For training $\phi''$, we assumed the availability of a dataset $\mathcal{D} = \{(\mathbf{x}_{0}, \mathbf{y}, \mathbf{m}, c)\}$. Besides the target video $\mathbf{x}_{0}$ and the reference conditioning image $\mathbf{y}$, we require both \textit{low-level} and \textit{high-level} conditioning for physical dynamics. At the low level, we require the aggregated velocity masks $\mathbf{m}$ defined in Section~\ref{sec:motion-control}. At the high level, instead, we associate each video with a textual description $c$ summarizing the effects of physical interactions. Since collecting real-world videos with such annotations is impractical, we generate synthetic data in Blender. Importantly, such simulated data still allows to generalize to real scenes, as we empirically verify in our experiments. We render scenes with boxes and cylinders placed on textured surfaces, and assign to each controlled object an initial velocity with random direction and magnitude. This procedure yields $\mathbf{x}_{0}$ as the rendered video, $\mathbf{y}$ as the first frame of the sequence (used for image-to-video conditioning), and $\mathbf{m}$ as the stack of per-frame velocity masks, which provide supervision of motion. To obtain the high-level descriptions $c$, we instead process each rendered video with a vision--language model (VLM), prompted to provide detailed video captions with particular focus on object interactions. The full prompt used for caption generation is reported in Appendix.

\vspace{-10px}\subsubsection{Inference.}
We assume as input an unseen image $\mathbf{y}$. An object mask is obtained for the target object using SAM2~\cite{ravi2024sam}, while the desired object velocity at the first frame is assumed to be provided by the user. We use this information to construct $\mathbf{m}_\odot$. We also prompt GPT-5~\cite{openai2025gpt5systemcard} for a description $c_\text{infer}$ of the effects on the scene if the object starts moving in the direction indicated by the user. The full prompt is in Appendix. Combining those with random noise $\mathbf{x}_T\sim\mathcal{N}(0, 1)$, we construct the input tuple $\{\mathbf{x}_T, \mathbf{y}_\text{input}, \mathbf{m}_\odot, c_\text{infer}\}$ compatible with our VDM equipped with $\phi''$. Our inference pipeline is illustrated in Figure~\ref{fig:Architecure} (right). Please note that at inference the user simply draws an arrow, mapped to RGB masks automatically.\looseness=-1

\vspace{-8px}\section{Experiments}
We describe the experimental setup in Section~\ref{sec:setup}. Section~\ref{sec:baselines} presents our main results and compares KineMask to the state of the art. Finally, Section~\ref{sec:analysis} shows a comprehensive analysis with ablations for different components of our method. Further ablations are in Appendix.\looseness=-1

\vspace{-6pt}\subsection{Setup}\label{sec:setup}
\subsubsection{Datasets.}
We generate two synthetic datasets for training and evaluation. Following Section~\ref{sec:data}, we render cubes and cylinders with random colors, moving on textured backgrounds from AmbientCG~\cite{ambientcg}. The first \textit{\textbf{Interactions}} dataset contains objects moving in random directions and interacting with each other. We also construct a \textit{\textbf{Simple Motion}} dataset, where isolated objects are moving in random directions without collisions. For both, we generate 10,000 training and 100 test samples. Test videos use disjoint colors and textures compared to training to ensure diversity. We visualize samples of the generated data in Appendix. We also include a \textit{\textbf{Real World}} set of 50 images, collected from the web~\cite{unsplash} or generated with ChatGPT’s image generator~\cite{hurst2024gpt}, used to assess generalization to real scenes with complex objects. Note that unlike \emph{Simple Motion} and \emph{Interactions}, \emph{Real World} does not include ground-truth motion trajectories. We use Tarsier~\cite{wang2024tarsier} to extract captions $c$ for synthetic data.\looseness=-1

\begin{figure*}[t]
    \centering
    \includegraphics[width=\linewidth]{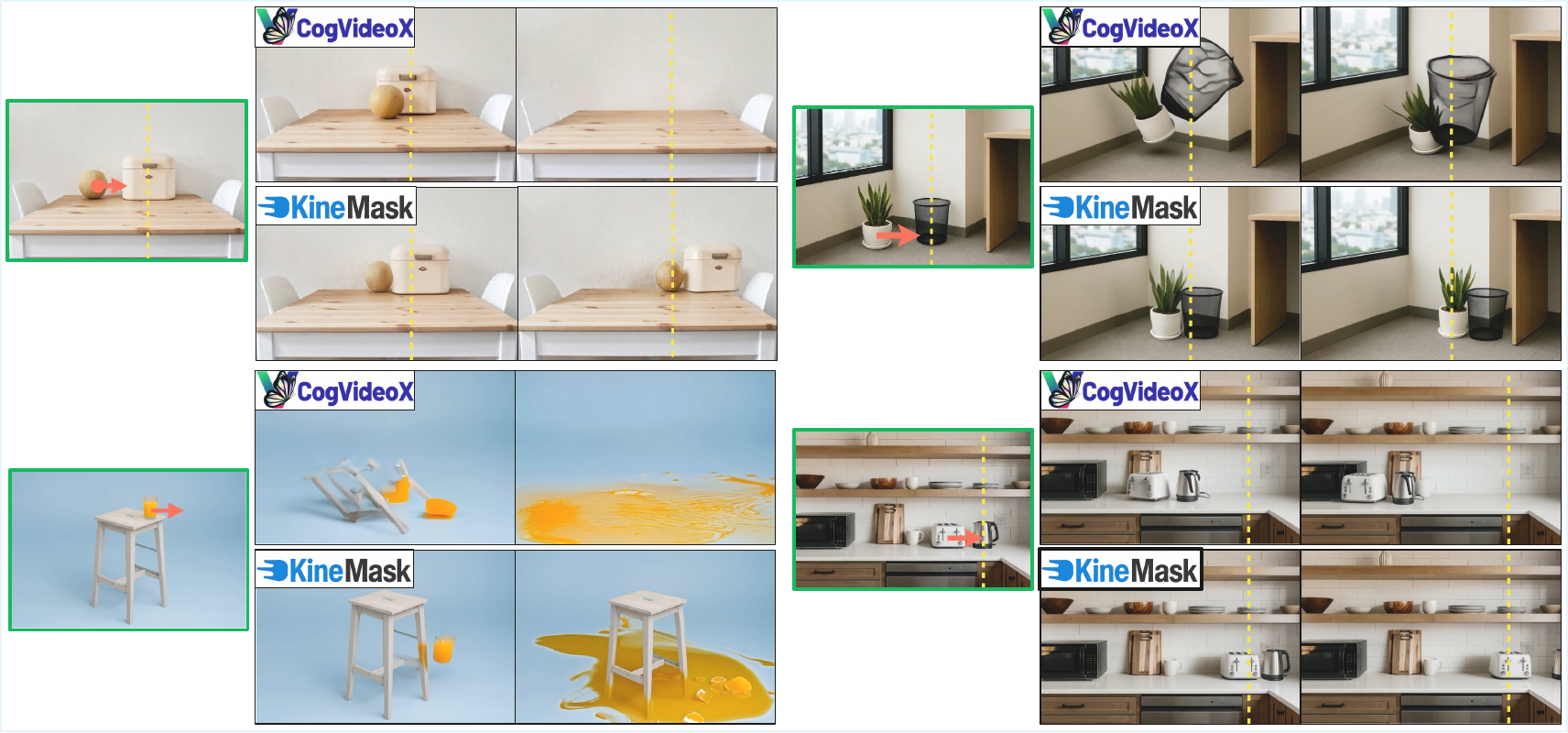}
    \caption{\textbf{Comparison with CogVideoX.} While CogVideoX often suffers from several failure modes, such as hallucinations and incorrect motions, \emph{KineMask} follows target motion and generates realistic object interactions. We improve object interactions in collisions and show causal effects of object motion.} \vspace{-12pt}
    \label{fig:Comparison}
\end{figure*}

\vspace{-10px}\subsubsection{Implementation.}
We adopt CogVideoX-I2V-5B (CogVideoX) as backbone for KineMask. We use ControlNet on the first 8 layers of the model, with weight at 0.5. For $\phi''$, we also apply a non-uniform sampling strategy for $f^{*}$, where frame selection is biased toward earlier frames, as rigid body interactions occur most frequently at the beginning of the simulated sequences. Further details are provided in Appendix. We generate 49 frames at inference.

\vspace{-10px}\subsubsection{Baselines.}
We consider two baselines using pretrained image-to-video models: CogVideoX~\cite{yang2024cogvideox} and Wan2.2-I2V-5B~\cite{wan2025wan} (Wan). We prompt both baselines with the same $c_\text{infer}$ as used for KineMask. Since we use CogVideoX as a backbone for KineMask, these two methods only differ by our training procedure. We choose TORA~\cite{zhang2025tora}, also based on CogVideoX, as a drag-based baseline. Since TORA requires the full drag trajectory, we draw it starting from the target object depending on the input velocity (see Appendix). We also include MotionI2V~\cite{shi2024motion} as mask-constrained drag method, that we prompt using SAMv2 masks and a drag trajectory similar to TORA. Finally, we evaluate Force Prompting (FP)~\cite{gillman2025force}, by mapping our input velocity to input force prompts~\cite{gillman2025force}. Force Prompting is also built on CogVideoX. More details are in Appendix.\looseness=-1

\vspace{-10px}\subsubsection{Metrics.}
For visual quality, we report the Fréchet Video Distance (FVD)~\cite{unterthiner2019fvd} and the mean squared error (MSE) between generated and ground-truth videos in our synthetic test sets. For motion, we compute the Fréchet Video Motion Distance (FVMD)~\cite{liu2024fr}, which isolates motion quality from appearance. Finally, we use SAM2~\cite{ravi2024sam} to extract semantic masks of objects in generated videos, and compute Intersection over Union (IoU) with ground-truth masks and assess geometric consistency.\looseness=-1

\begin{figure*}[t]
    \centering
        \centering
        \setlength{\tabcolsep}{0px}
        \resizebox{0.95\linewidth}{!}{
        \begin{tabular}{c c@{\hspace{4px}} c c@{\hspace{4px}} c c}
            \multicolumn{2}{c}{Direction} & \multicolumn{2}{c}{Speed} & \multicolumn{2}{c}{Object}\\
            \fcolorbox{darkgreen}{white}{\includegraphics[width=0.16\textwidth]{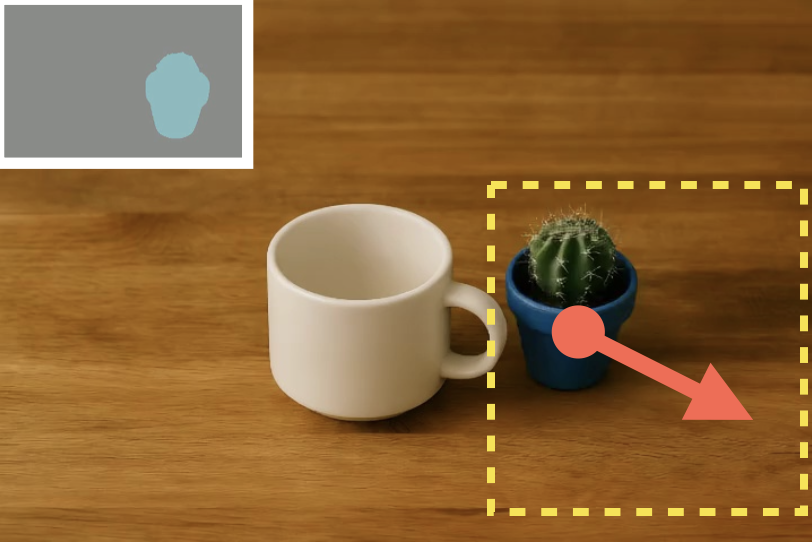}} &
            \includegraphics[width=0.16\textwidth]{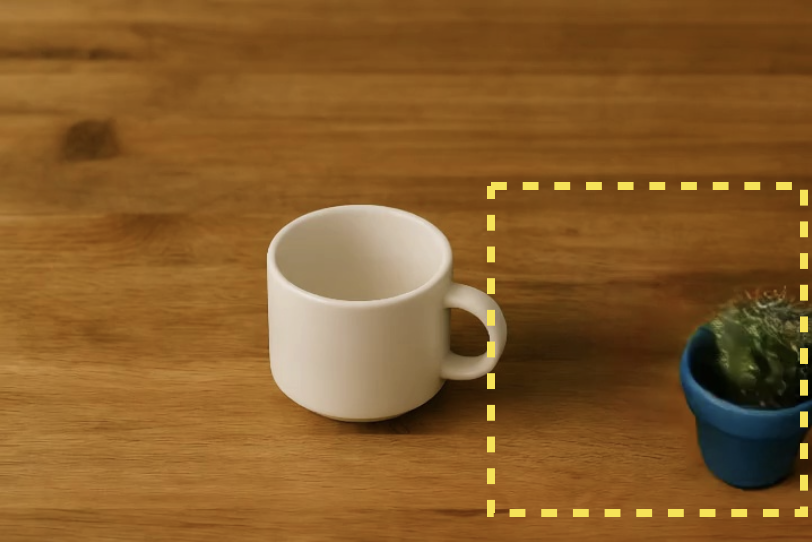} & 
            \fcolorbox{darkgreen}{white}{\includegraphics[width=0.16\textwidth]{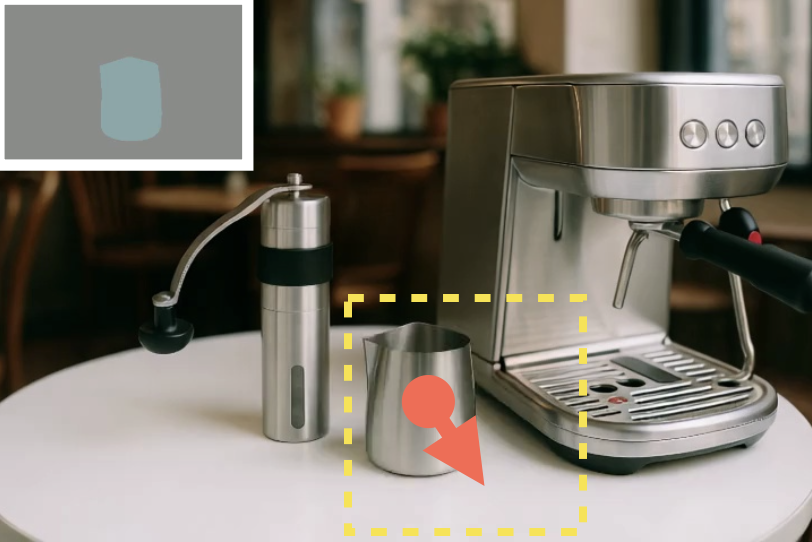}}&
            \includegraphics[width=0.16\textwidth]{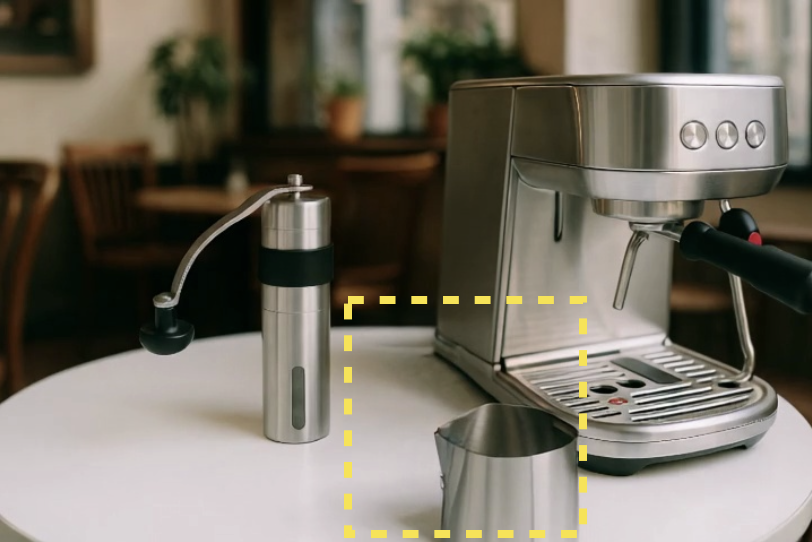} & 
            \fcolorbox{darkgreen}{white}{\includegraphics[width=0.16\textwidth]{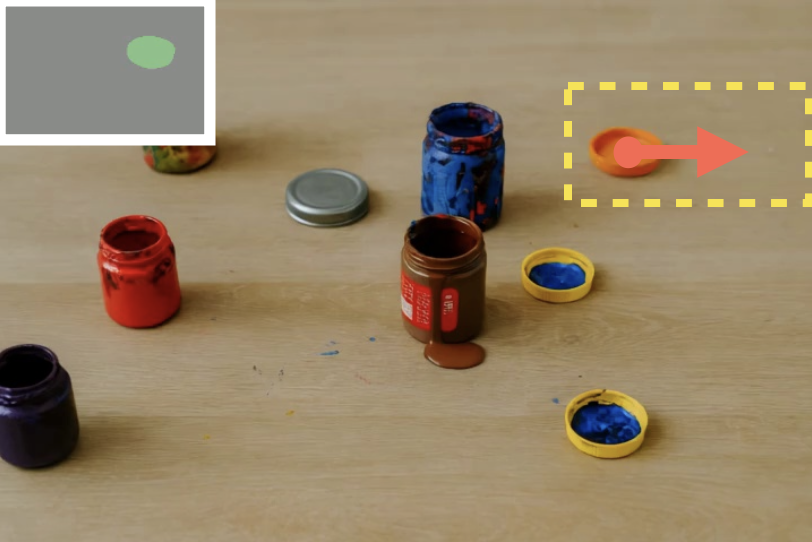}} &
            \includegraphics[width=0.16\textwidth]{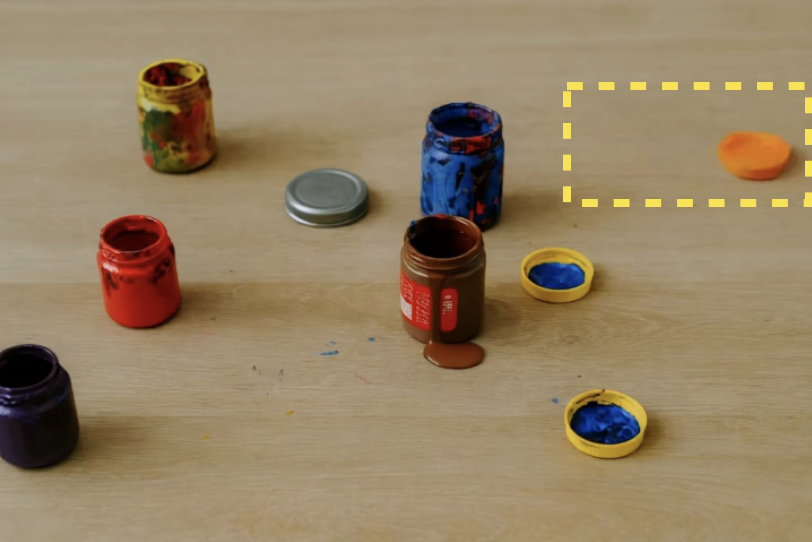} \\
            
            \fcolorbox{darkgreen}{white}{\includegraphics[width=0.16\textwidth]{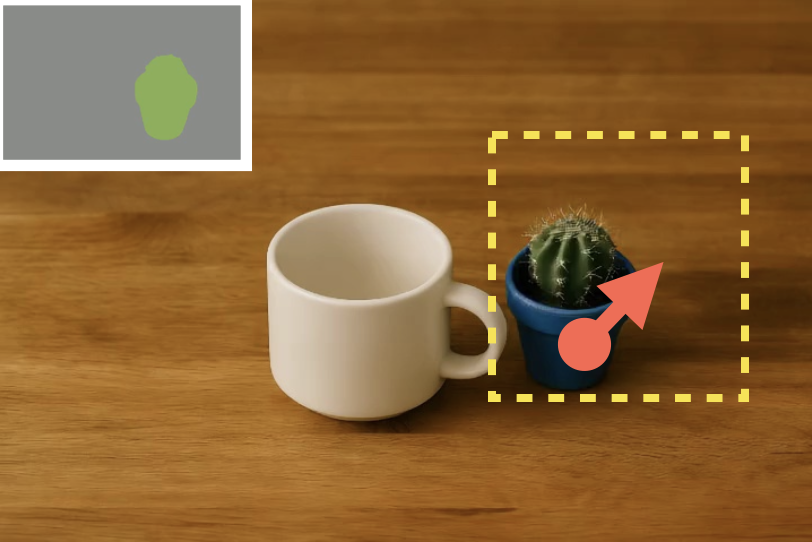}} &
            \includegraphics[width=0.16\textwidth]{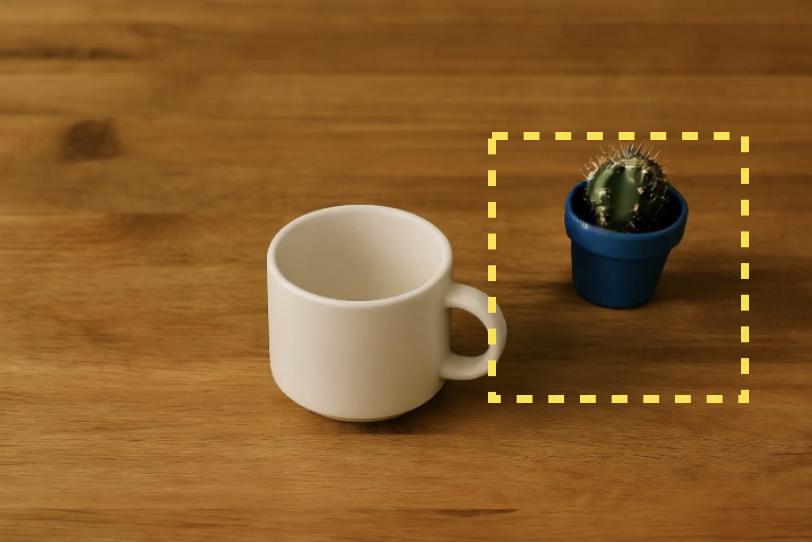} & 
            \fcolorbox{darkgreen}{white}{\includegraphics[width=0.16\textwidth]{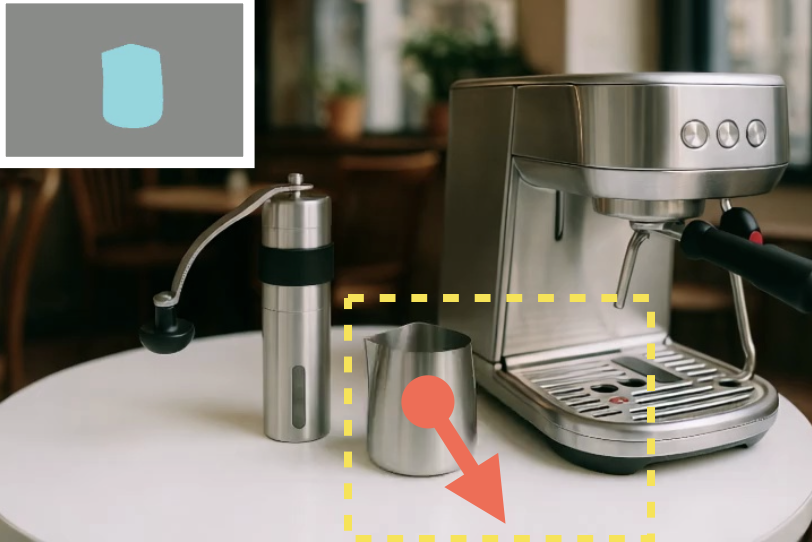}} &
            \includegraphics[width=0.16\textwidth]{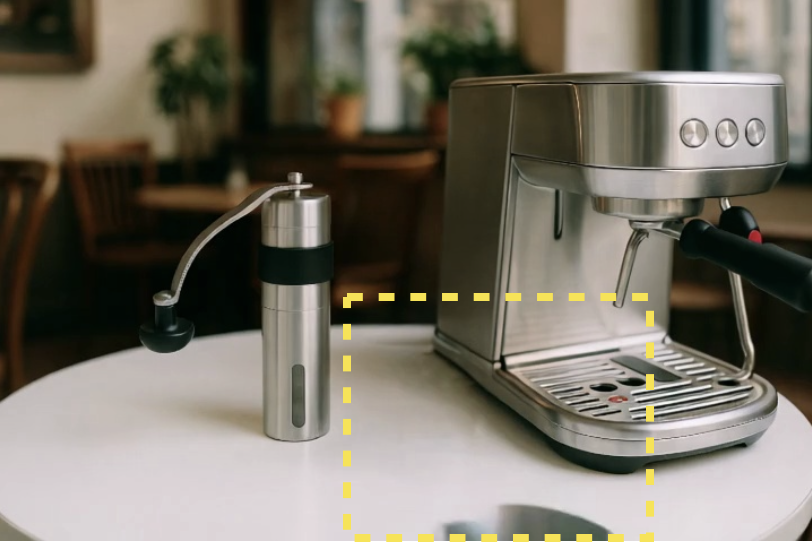} & 
            \fcolorbox{darkgreen}{white}{\includegraphics[width=0.16\textwidth]{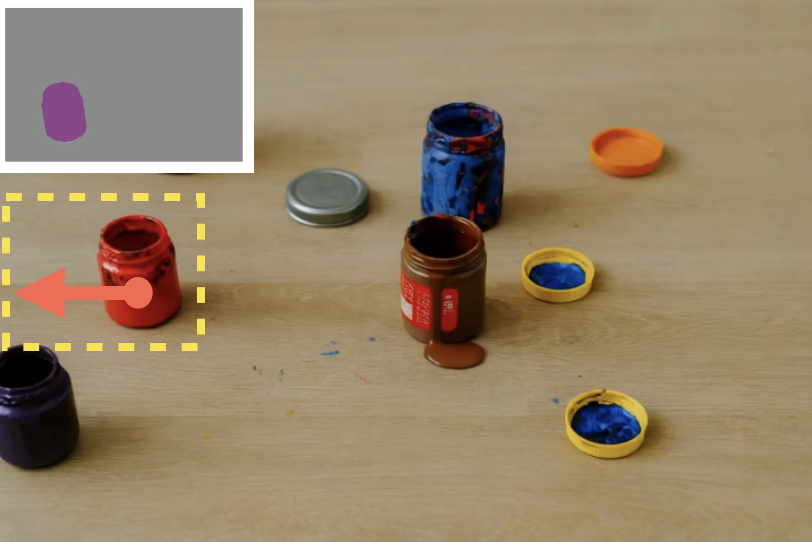}} &
            \includegraphics[width=0.16\textwidth]{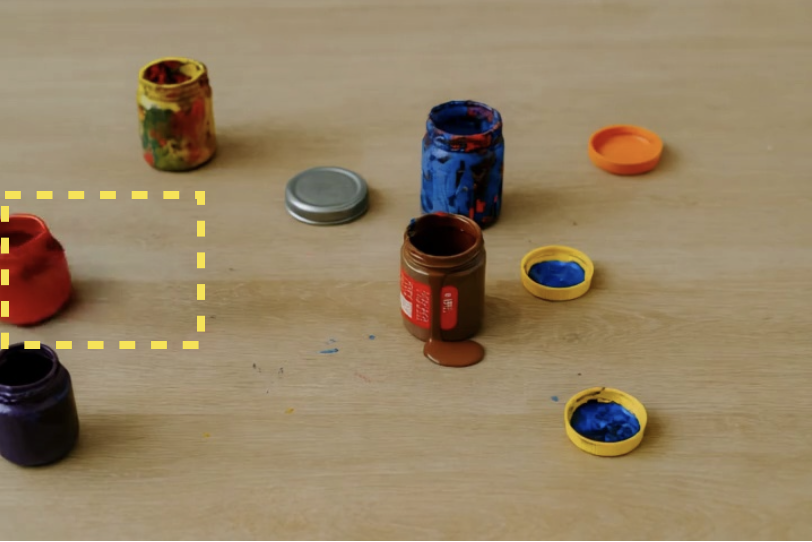}\\
        \end{tabular}
        }
        \caption{\textbf{Degrees of freedom.} We show control of different aspects of KineMask outputs. We can choose different directions (left), speed (middle), and objects to move (right), opening potential for world modeling.}
        \label{fig:qual-simple-motion}
        \vspace{-10pt}
\end{figure*}

\vspace{-5pt}\subsection{Comparison with baselines}\label{sec:baselines}
\subsubsection{Qualitative Comparison.}
We demonstrate qualitative improvements on \emph{Real World} scenes. While for space reasons we compare only with the backbone CogVideoX in Figure~\ref{fig:Comparison}, we include comparisons against all baselines in the Appendix. We noticed that CogVideoX suffers from unrealistic interactions, making objects fly or disappear. Instead, KineMask generates realistic interactions with other objects if they are present in the path of motion of the initial moving object, showing a correct understanding of rigid body dynamics. We also show (Figure~\ref{fig:Comparison}, bottom left) complex interactions that require implicit 3D understanding, making the glass of juice fall and crash as a result of motion. Our realistic real world outputs show a strong generalization of the knowledge acquired from simulated videos.\looseness=-1

\vspace{-10pt}\subsubsection{User Study.}
\begin{wrapfigure}{r}{0.55\textwidth} 
    \vspace{-25pt}
    \centering
    \includegraphics[width=\linewidth]{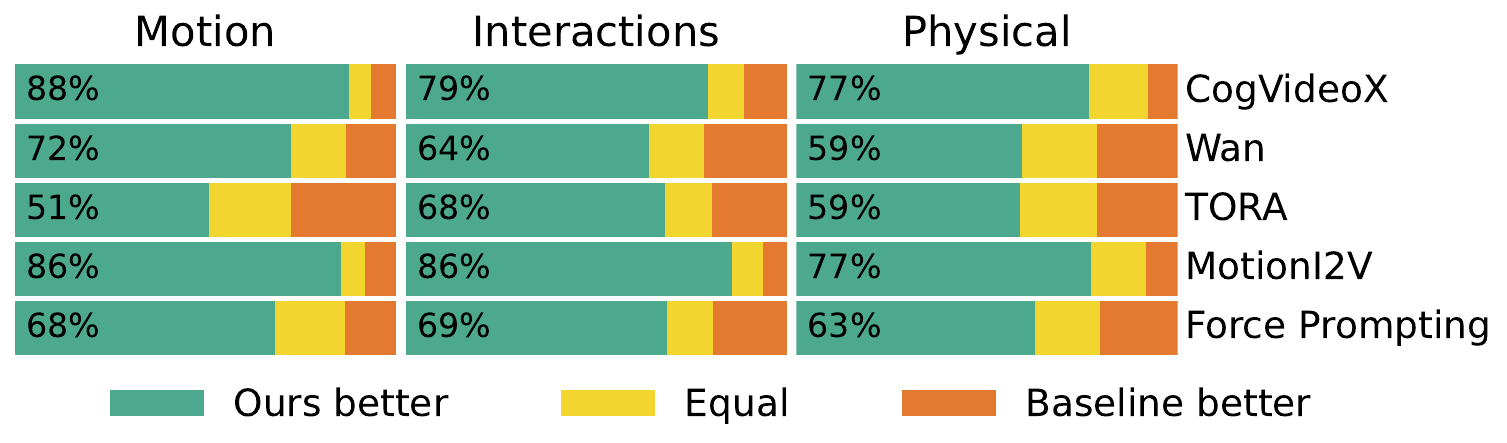}
    \caption{We widely outperform baselines on motion fidelity, interaction quality, and overall physical consistency.}\vspace{-20px}
    \label{fig:userstudy}
\end{wrapfigure}
In Figure~\ref{fig:userstudy}, we show results of a user study with 30 participants, who perform pairwise comparisons of videos generated by KineMask and baselines on \emph{Real World}. Participants are presented with initial images and input object velocities indicated by arrows. We ask them to evaluate outputs on three different axes: motion fidelity to the control signal (Motion), realism of the object interactions (Interactions), overall physical consistency (Physical). Each answer is collected with a three-answer forced choice format, where we compare our outputs with baselines and we ask for user preferences, allowing also to reply ``both have the same quality''. The full details are presented in Appendix.
The collected user preferences indicate that \textit{our method significantly outperforms  all baselines in all questions}. This demonstrates that existing methods are unsuitable for rendering interactions with starting conditions only. Kinemask, instead, renders realistic interactions.\looseness=-1

\vspace{-7pt}\subsection{Analysis and ablation studies}\label{sec:analysis}
\subsubsection{Low-level motion control.}\label{sec:simple}
In our first set of experiments, we evaluate KineMask on low-level motion control. To do so, we train KineMask on \emph{Simple Motion} with a prompt $c_\emptyset=$``\textit{An object moving on a surface}", hence discarding the rich description $c$ extracted by Tarsier. At inference, we assume $c_\text{infer}=c_\emptyset$. Doing so, we isolate the effects of low-level conditioning from high-level textual control, allowing for a fair assessment of KineMask as a motion conditioning method.\looseness=-1

\vspace{-5pt}\paragraph{Fine-grained control.} In Figure~\ref{fig:qual-simple-motion}, we present results on different degrees of freedom of KineMask. Despite being trained on basic synthetic data, KineMask generalizes motion control to complex real-world scenes, coherently with the findings in~\cite{gillman2025force}. In particular, we show that KineMask achieves disentangled control over different directions, speed, and objects. This allows for a fine-grained evaluation of motion dynamics, precious for world modeling.

\begin{figure*}[t]
    \centering
    \includegraphics[width=\linewidth]{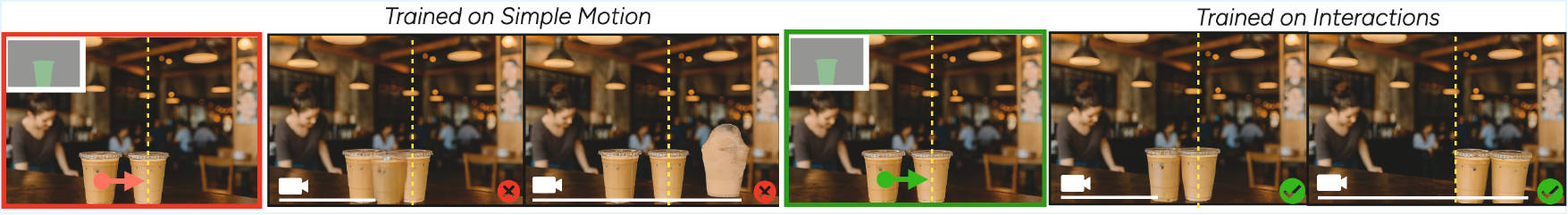}
    \caption{\textbf{Impact of training data.} While \emph{KineMask} trained on \emph{Simple Motion} is able to generalize to \emph{Real World} images, the lack of object interactions in \emph{Simple Motion} results in hallucinations (top). Training on \emph{Interactions} results in collisions and plausible motion of pushed objects (bottom).}
    \label{fig:qual-data-impact}\vspace{-12pt}
\end{figure*}

\vspace{-5pt}\paragraph{Two-stage training.}
\begin{wraptable}{r}{0.52\textwidth} 
        \vspace{-20pt}
        \centering
        \setlength{\tabcolsep}{4pt}
        \renewcommand{\arraystretch}{1.2}
        \resizebox{0.9\linewidth}{!}{
        \begin{tabular}{l|cccc}
        
            \multicolumn{1}{c}{}& \multicolumn{4}{c}{\textit{Simple Motion}} \\
            \toprule
            \textbf{Method} & \textbf{MSE}$\downarrow$ & \textbf{FVD}$\downarrow$ & \textbf{FMVD}$\downarrow$ & \textbf{IoU}$\uparrow$ \\
            \midrule
            CogVideoX    & 158.3 & 601.1 & 1504.6 & 0.051 \\
            2nd stage only & 86.3 & 288.8 & 201.0  & 0.237 \\
            Ours \small{(1st + 2nd stage)} & \textbf{47.2} & \textbf{160.3} & \textbf{199.8}  & \textbf{0.367} \\\midrule
            \color{gray}{1st stage w/ masks} & \color{gray}{24.9} & \color{gray}{89.7} & \color{gray}{165.6} & \color{gray}{0.684}  \\      
            \bottomrule
        \end{tabular}}
    \caption{\textbf{Ablation on training}.  We show that our two-stage training strategy considerably boosts results and approaches an upper bound exploiting privileged information (1st stage w/ masks).}
    \label{tab:simple-motion}
    \vspace{-20pt}
\end{wraptable}
Table~\ref{tab:simple-motion} shows the importance of our two-stage training introduced in Section~\ref{sec:motion-control}. We test KineMask on \emph{Simple Motion} only, to study the effects of motion control in isolation. We report results of CogVideoX prompted with $c$ as a lower bound. As baseline, we train directly $\phi''$ (2nd stage only), without first-stage pretraining. As an upper bound, we also evaluate $\phi'$ using ground truth object masks provided for every frame at test time (1st stage w/ masks). Our two-stage strategy boosts considerably all metrics compared to training $\phi''$ directly, approaching the upper bound. We noticed that the IoU is very sensitive to minor displacement errors of the moving objects (0.367 vs 0.684), that do not impact the quality of motion as captured by the other metrics.

\vspace{-7px}\subsubsection{Impact of data.}\label{sec:interactions}
Next, we explore the effects of training data on KineMask for generating realistic rigid body interactions. We compare models trained on synthetic samples with and without interactions, and draw conclusions on the capability of VDMs to learn motion. We still use $c_\emptyset$ at both training and inference, as in Section~\ref{sec:simple} (Low level motion control). 

\vspace{-5pt}

\paragraph{Data influence.}
We evaluate KineMask's capability to synthesize object interactions depending on the training data. We first test the model trained on \emph{Simple Motion} and apply it to \emph{Real World} data. We deliberately set input velocities to directions that should generate collisions among objects in the scene. The VDM fails to render realistic collisions (Figure~\ref{fig:qual-data-impact}, left). However, \emph{KineMask} trained on the \emph{Interactions} set produces accurate interactions for the same input velocities (Figure ~\ref{fig:qual-data-impact}, right). Hence, \textit{\emph{KineMask} trained on appropriate data allows to render complex object interactions}. Moreover, this experiment yields important insights: while VDMs are robust to visual distribution shift, as even synthetic training data do not influence the realism of \emph{Real World} generated scenes, their preservation of motion understanding seems to be more impacted, since training on \emph{Simple Motion} prevents to generate interactions. 

\begin{figure*}[t]
    \centering
    \includegraphics[width=\linewidth]{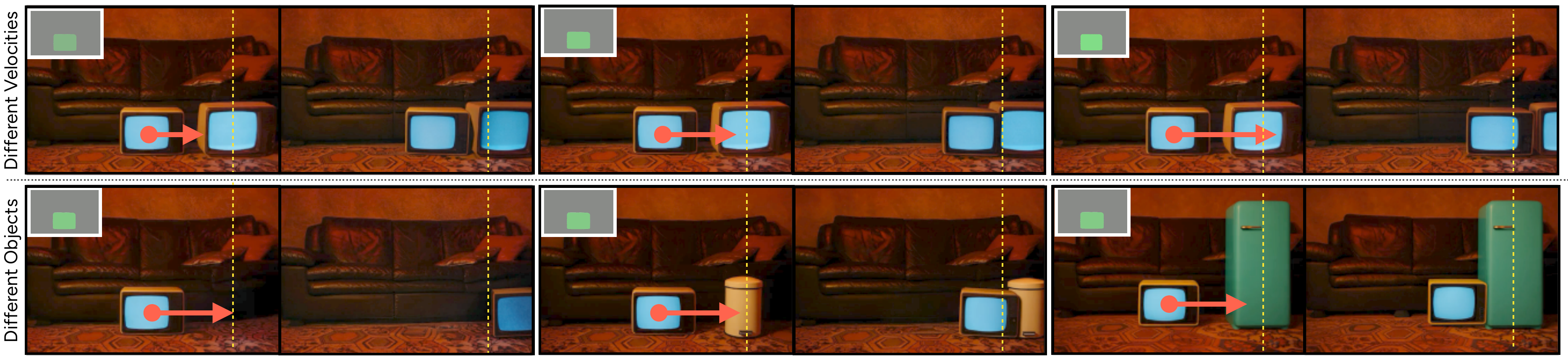}
    \caption{On top, we show how different velocities impact motion effects, suggesting causal understanding. At the bottom, we show that the environment impacts results, for the same velocity but different objects, suggesting mass awareness.}\vspace{-10pt}
    \label{fig:causality}
\end{figure*}

\vspace{-3pt}\paragraph{Quantitative evaluation.}

\begin{wraptable}{r}{0.52\textwidth}
\vspace{-25pt}
\centering
\setlength{\tabcolsep}{4pt}
\resizebox{\linewidth}{!}{%
\begin{tabular}{ll|cccc}
  \multicolumn{2}{c}{} & \multicolumn{4}{c}{\emph{Interactions}} \\
  \toprule
  \textbf{Method} & \textbf{Training} & \textbf{MSE}$\downarrow$ & \textbf{FVD}$\downarrow$ & \textbf{FMVD}$\downarrow$ & \textbf{IoU}$\uparrow$ \\
  \midrule
  CogVideoX & - & 344.6 & 807.3 & 3514.9 & 0.192 \\
  CogVideoX & Interactions & 201.9 & 368.2 & 162.9 & 0.243 \\
  \midrule
  KineMask & \emph{Simple Motion} & 166.2 & 301.1 & 160.5 & 0.334 \\
  KineMask & \emph{Interactions} & \textbf{158.7} & \textbf{250.7} & \textbf{143.8} & \textbf{0.355} \\
  \bottomrule
\end{tabular}%
}
\caption{\textbf{Effects of data.} Compared with models trained on \emph{Simple Motion}, training on \emph{Interactions} data considerably boosts performance in all metrics.}
\label{tab:interaction_metrics}
\vspace{-20pt}
\end{wraptable}

We validate the importance of specific data with a quantitative evaluation on \emph{Interactions}, following the same protocol as in Section~\ref{sec:simple} (Low-level motion control). As shown in Table~\ref{tab:interaction_metrics}, KineMask achieves the best results when trained on the \emph{Interactions} dataset. Notably, even when trained only on \emph{Simple Motion}, it greatly outperforms CogVideoX Zero-Shot and CogVideoX finetuned on \emph{Interactions}, highlighting the need of our research to improve object interactions.\looseness=-1

\vspace{-5pt}\paragraph{Emergence of causality.} Figure~\ref{fig:causality} shows results of KineMask trained on \emph{Interactions} and tested with different inputs on a \emph{Real World} scene with interacting objects. On top, as velocity increases, the interactions also change, indicating that the model captures causality of collisions. At the bottom, \textit{with the same velocity}, we inpaint the input image, removing the interacting object, or replacing it with objects of different mass. The different final position of the moving object demonstrates understanding of mass in collisions. Note that this result is challenging for drag-based methods since they imply prior knowledge of the object final position. KineMask instead maps different target positions to the same input, depending on the environment.\looseness=-1

\begin{figure*}[t]
    \centering
    \includegraphics[width=\linewidth]{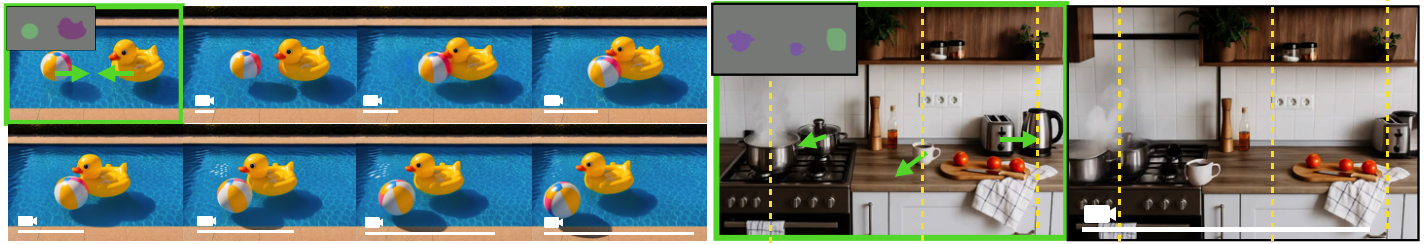}
    \vspace{-15pt}
    \caption{\textbf{Multi-Object Control.} Generation examples of \emph{KineMask} for multiple object control and interactions, using velocity masks for different objects as input generates objects colliding with each other or with other objects in the scene.} \vspace{-12pt}
    \label{fig:multi}
\end{figure*}

\vspace{-8pt}\paragraph{Multi-Object Control}
In Figure~\ref{fig:multi} we show examples of multi-object control, and interactions. We set initial velocity masks for different objects, and show that KineMask can render motion and interactions for multiple objects at the same time. Left, shows two objects moving towards each other, while right shows multiple interactions with 3 objects in a kitchen.
As visible, we preserve realism of interactions if both objects collide with each other or with other objects.\looseness=-1

\begin{figure*}[t]
    \centering
    \includegraphics[width=\linewidth]{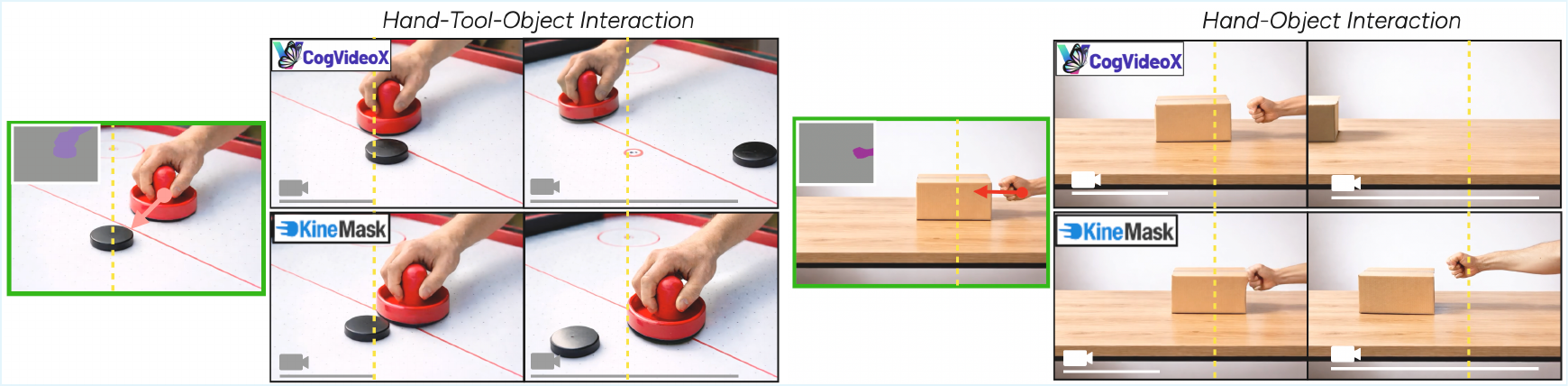}
    \caption{\textbf{Generalization.} \emph{KineMask} trained on \emph{Interactions} is also able to generalize its knowledge to \emph{Hand-Tool-Object} (Left) and \emph{Hand-Object} (Right) interactions, showing plausible generations that respect causality.}
    \label{fig:human-object}\vspace{1pt}
\end{figure*}

\begin{figure*}[t]
    \centering
    \includegraphics[width=\linewidth]{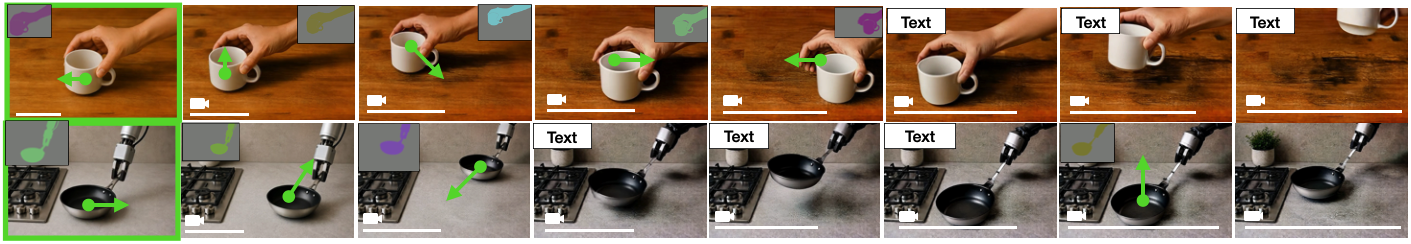}
    \vspace{-15pt}
    \caption{\textbf{Sequential Generation.} Generation examples applying KineMask sequentially, by combining our velocity masks and text only.} \vspace{-12pt}
    \label{fig:practicalitymain}
\end{figure*}

\vspace{-5pt}\paragraph{Generalization.} By training on interactions \emph{KineMask} is able to generalize its knowledge not only to object-object interaction but also to hand-object and hand-tool-object interaction as it is shown in Figure~\ref{fig:human-object}. By exploiting the knowledge gained in our two stage training with synthetic data to learn plausible collisions, and its powerful prior knowledge from the real world, \emph{KineMask} is able to generate plausible interactions. Notice that CogVideoX generates unrealistic motion, where the objects start to move in a certain direction without any touch from the hand-tool. While KineMask follows the input given by the user, respects causality and generates plausible interactions. Furthermore, our work focuses on understanding causality and rigid body interactions to study potential outcomes of actions in the world, which can be applicable to fields like robotics. Figure ~\ref{fig:practicalitymain} show a sequential generation using an image of hands or a robotic gripper manipulating objects, in which we apply KineMask sequentially. We also propose actions using text, to showcase the preservation of textual control. \looseness=-1

\vspace{-7px}\subsubsection{High-level text conditioning.}\label{sec:text}
We now enable the usage of $c$ at training and $c_\text{infer}$ at inference. This allows us to assess the effects of text introduced in KineMask, and to draw insights on generalization.\looseness=-1

\begin{figure*}[t]
    \centering
    \includegraphics[width=\linewidth]{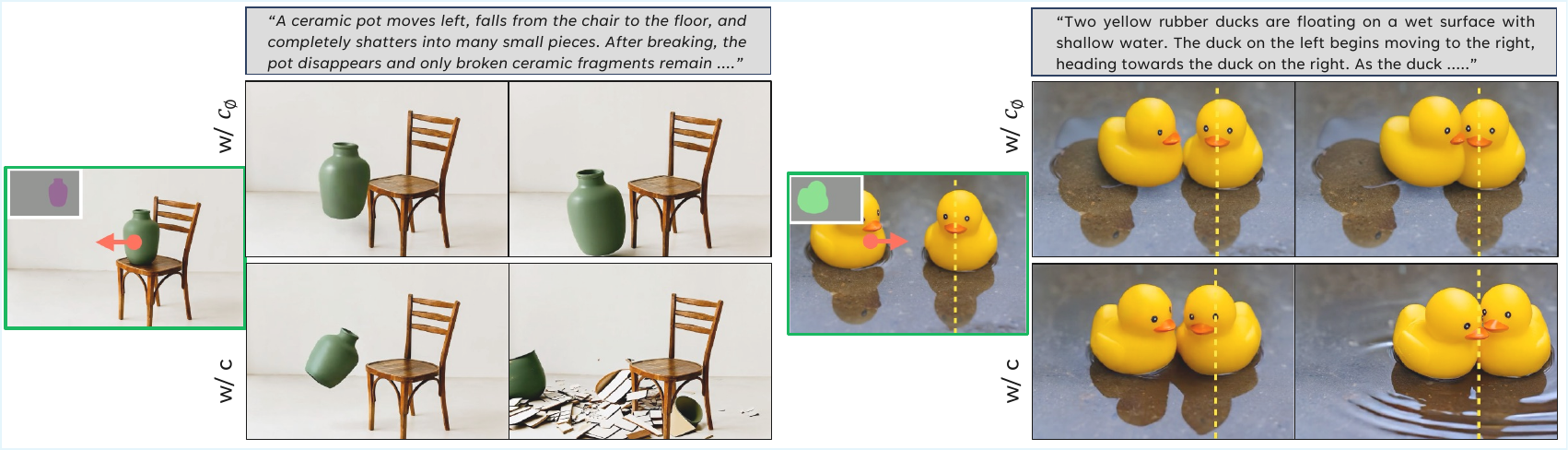}
    \vspace{-15px}
    \caption{\textbf{Impact of text.} Training with rich captions $c$ makes KineMask generate effects that go beyond the synthetic data used for training, exploiting the prior knowledge of the VDM. While we prompt both models with $c_\text{infer}$, KineMask trained with $c$ (bottom row) is able to generate complex effects of the interactions, such as crashing objects, and water effects. Full prompts in Appendix.}
    \label{fig:3d-impact}
    \vspace{-10px}
\end{figure*}

\paragraph{Qualitative evaluation.}
We compare KineMask trained with $c_\emptyset$ against training with $c$, while using $c_\text{infer}$ for inference in both cases. Doing so, we aim to evaluate the effects of textual prompts describing interactions during training. As shown in Figure~\ref{fig:3d-impact}, training with rich captions $c$ allows the method to render interactions beyond those used for training, successfully exploiting the VDM prior knowledge in the rendered videos. Indeed, the model trained with $c_\emptyset$ fails to follow some effects in $c_\text{infer}$, such as vase breaking, or small waves forming. Instead, the model trained with $c$ is able to benefit from additional information described in $c_\text{infer}$, correctly breaking the vase and perturbing the water. Please also note that the captions used in training are always tied to synthetic elements in \emph{Interactions}, and therefore have limited diversity. Nevertheless, this still enables successful transfer to $c_\text{infer}$ prompts describing complex effects that go beyond the training distribution of KineMask.\looseness=-1

\vspace{-3px}\paragraph{Quantitative evaluation.} 
\begin{wraptable}{r}{0.48\textwidth} 
\vspace{-25pt} 
\centering
\small
\setlength{\tabcolsep}{2pt}

\resizebox{\linewidth}{!}{%
\begin{tabular}{cc|cccc}
  \multicolumn{2}{c}{} & \multicolumn{4}{c}{\emph{Interactions}} \\
  \toprule
  \textbf{Train} & \textbf{Infer} & \textbf{MSE}$\downarrow$ & \textbf{FVD}$\downarrow$ & \textbf{FMVD}$\downarrow$ & \textbf{IoU}$\uparrow$ \\
  \midrule
  $c_\emptyset$ & $c_\emptyset$    & \textbf{158.7} & 250.7 & \textbf{143.8} & 0.355 \\
  $c_\emptyset$ & $c_\text{infer}$ & 174.4 & 238.8 & 161.3 & 0.356 \\
  $c$           & $c_\text{infer}$ & 160.9 & \textbf{231.3} & 174.4 & \textbf{0.376} \\
  \bottomrule
\end{tabular}%
}
\caption{\textbf{Ablation on prompts.} Training with $c$ improve object consistency and quality on synthetic data.}\looseness=-1
\label{tab:text-quantitative}
\vspace{-20px}
\end{wraptable}
We evaluate metrics with different textual conditioning in Table~\ref{tab:text-quantitative}. We consider trainings with $c_\emptyset$, using $c_\emptyset$ or $c_\text{infer}$ at inference. Results demonstrate that using $c$ at training time boosts object consistency (IoU 0.376 vs 0.356 of the second best configuration), overall realism (FVD 231.3 vs 238.8, MSE on par), while we lose some motion consistency (FMVD 174.4 vs 143.8 of the best model). However, note that Table~\ref{tab:text-quantitative} includes \textit{only evaluation on synthetic data}, where we have access to a motion ground truth. Considering the improved qualitatives on \emph{Real World} and the benefit on complex interactions (Figure~\ref{fig:3d-impact}) we believe the use of captions $c$ during training is still beneficial for the final model to improve robustness and preserve its capabilities.\looseness=-1

\vspace{-10px}\subsubsection{Generalization to different models.}\label{sec:text}
\begin{wrapfigure}{r}{0.46\textwidth} 
    \vspace{-1pt}
    \centering
    \includegraphics[width=\linewidth]{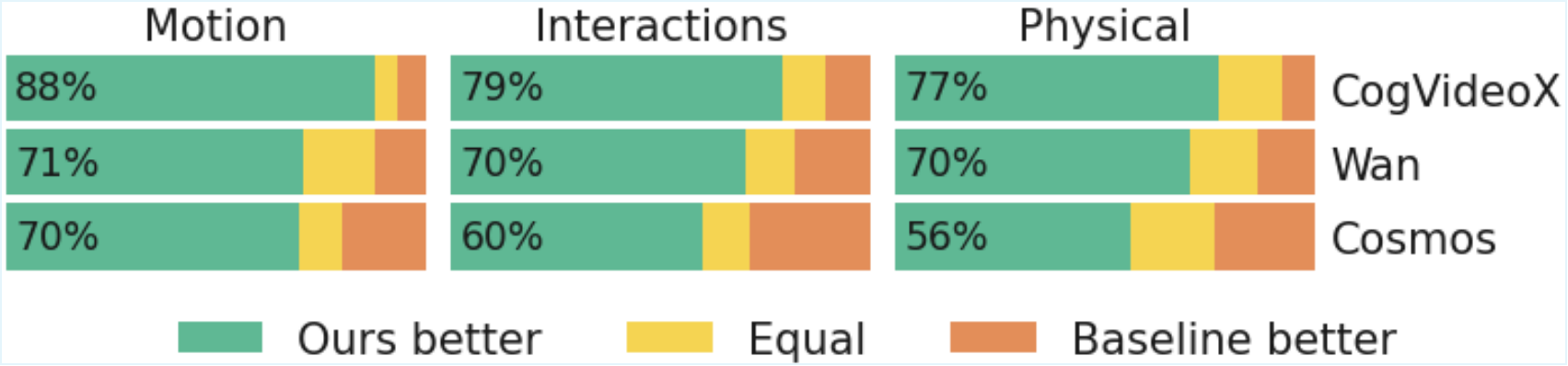}
    \caption{All versions of \emph{KineMask} outperform the original models, on motion fidelity, interaction quality and overall physical consistency. Showing its applicability, and generalization, bringing improvements to different video diffusion models.}\vspace{-20px}
    \label{fig:extra_userstudy}
    \vspace{-1pt}
\end{wrapfigure}
\begin{figure*}[t]
    \centering
    \includegraphics[width=\linewidth]{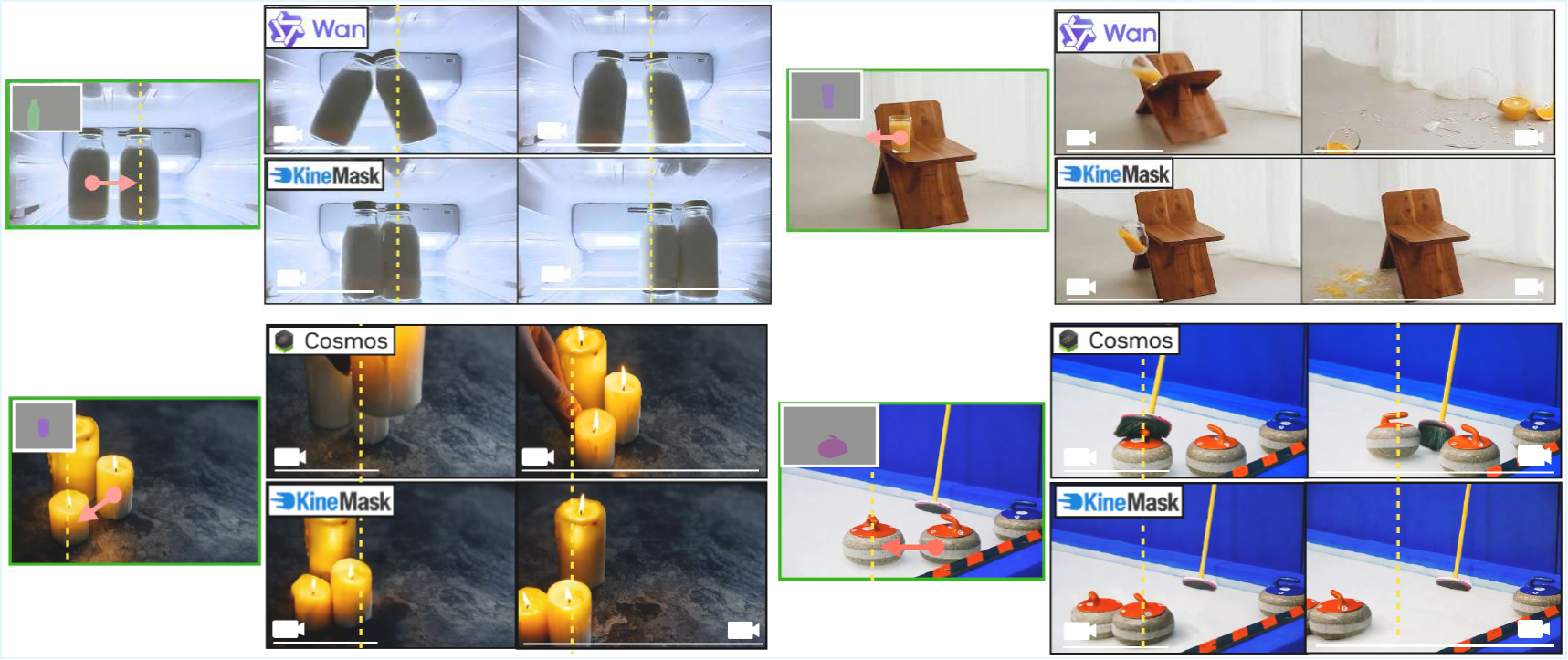}
    \vspace{-15pt}
    \caption{\textbf{Qualitative comparison with Wan and Cosmos.} \emph{KineMask-Wan} and \emph{KineMask-Cosmos} follows target motion and generates realistic object interactions, while the original backbones show implausible motions. This demonstrates that apart from our original implementation on top of CogVideoX, KineMask generalize to different VDM's bringing improvements to all models.} \vspace{-12pt}
    \label{fig:add_models}
\end{figure*}

Beyond CogVideoX, the \emph{KineMask} approach generalizes to other VDM's. To verify the generalization, we test \emph{KineMask} in combination with Wan2.2-5B~\cite{wan2025wan} and Cosmos2.5-2B~\cite{agarwal2025cosmos}. Following the same protocol as in ~\ref{sec:baselines}, we perform an additional human study (Figure~\ref{fig:extra_userstudy}) to evaluate the generations of \emph{KineMask-Wan}, \emph{KineMask-Cosmos} and compare with the respective backbone models for each case. The collected responses indicate that all versions of \emph{\mbox{KineMask} outperform the original models}, showing better motion, interactions, and physical realism. Additionally, Figure~\ref{fig:add_models} top row shows the comparison of Wan vs KineMask-Wan, at the bottom we show Cosmos vs KineMask-Cosmos. In both cases the implementation of \emph{KineMask} improves the original generation of each model respectively, synthesizing better object interactions, showing plausible collisions with liquid or flame movement, and glass breaking as a result of a fall, contrary to the original backbones that show hallucinations in motion and interactions. As expected, we saw that the implementation of KineMask in models that have better physical prior knowledge about the real world leads to more realistic generations with \emph{KineMask}, showing better physical effects created from interactions. \looseness=-1

\vspace{-5px}\section{Conclusion}\label{sec:discussion}

We introduced \emph{KineMask}, a framework to enable video generation of object interactions and effects in complex scenes. Our experiments show that \emph{KineMask} achieves significant improvements over state-of-the-art models of comparable size, synthesizing realistic multi-object interactions while allowing control over variable object velocities, and demonstrating its applicability and generalization to different VDMs. We believe this methodology can inform future work on world models, with potential implications for robotic manipulation, planning, and embodied decision making.\looseness=-1

\vspace{-3pt}\paragraph{Limitations and future work.} While effective, KineMask’s low-level conditioning is limited to velocity, whereas real-world motion depends on factors such as friction, mass, shape and air resistance. Moreover, we believe that extending to soft-body interactions will further improve world modeling capabilities. As shown in Section~\ref{sec:text}, combining text-based conditioning with KineMask enhances realism, highlighting the complementary roles of high-level textual guidance and low-level video control. These results support the joint use of both modalities for physically grounded world modeling, and we speculate that advances in multimodal language models~\cite{shukor2025scaling,team2023gemini} may enable text-based physical reasoning to complement video generation. KineMask is an initial study of object-centric rigid-body control, focusing on center-of-mass  motion rather than part-level actuation, this is not represented in our training data, but they could in principle be simulated and used to build richer control, making them a direction for future work.  On the other hand, KineMask is trained with static cameras, and does not explicitly render or control camera motion. However, camera motion could also be simulated, making camera-motion also a direction for future work.\looseness=-1



%
%
\vspace{-3pt}\subsubsection{Acknowledgement.} This work was supported by the Institute of Information \& Communications Technology Planning \& Evaluation (IITP) grant funded by the Korean Government (MSIT) (No. RS-2024-00457882, National AI Research Lab Project).
\bibliographystyle{splncs04}
\bibliography{main}

\clearpage 
\section{Appendix}


\subsection{Additional details}\label{sec:appendix-details}

\subsubsection{Used prompts.} 
\begin{figure*}[t]
    \centering
    \resizebox{\linewidth}{!}{
    
    \includegraphics[width=0.19\linewidth]{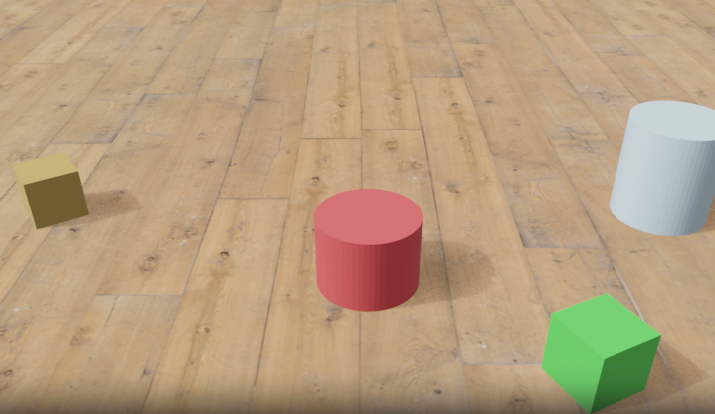}
    \hfill
    \includegraphics[width=0.19\linewidth]{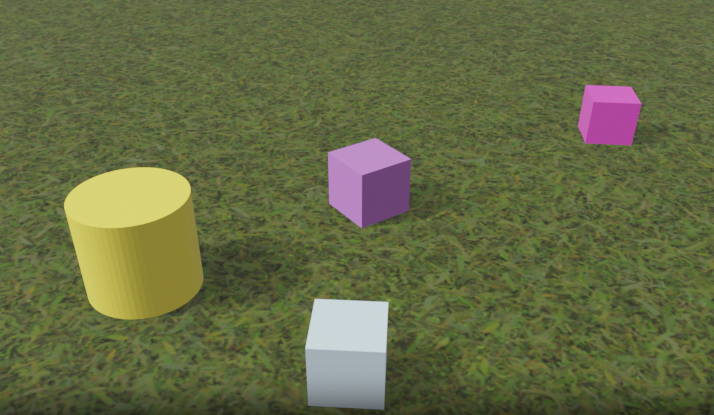}
    \hfill
    \includegraphics[width=0.19\linewidth]{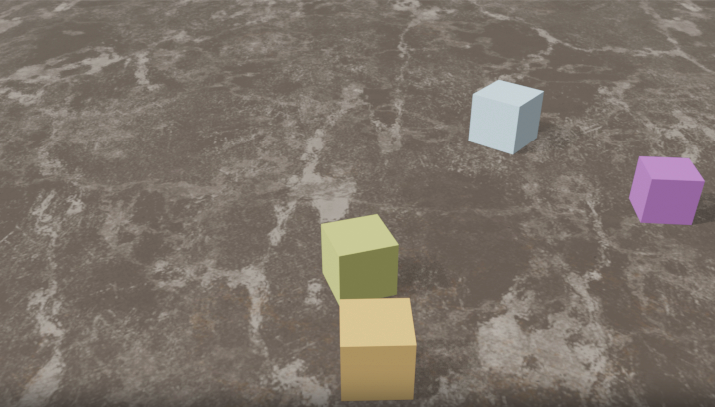}
    \hfill
    \includegraphics[width=0.19\linewidth]{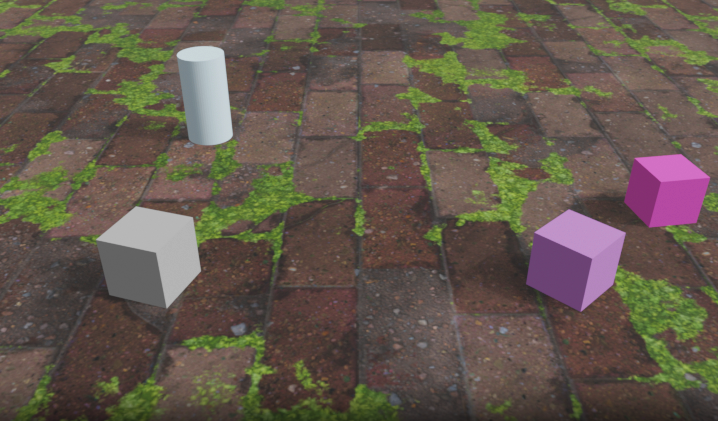}
    \hfill
    \includegraphics[width=0.19\linewidth]{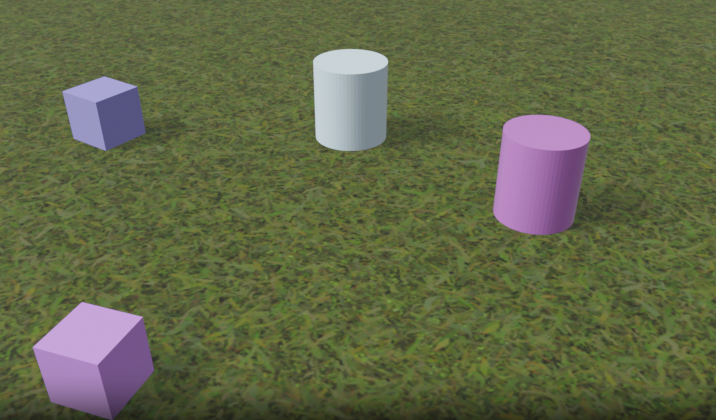}
    \hfill
    }
    \caption{\textbf{In-context learning examples.} Providing examples to GPT helps the generation of the text $c_\text{infer}$ following a desired format.}
    \label{fig:icl}
    \vspace{-5pt}
\end{figure*}

We report here the prompt used for $c$ generation at training time and $c_\text{infer}$ used for inference time. We prompt Tarsier~\cite{wang2024tarsier} to extract $c$ with: \textit{``Describe the video in a concise way, covering all motions and collisions between objects.''}. This simple prompt is already resulting in sufficiently good descriptions, please note that we tested other current VLMs for this task, however with tarsier we got the best video captions for our synthetic blender videos. For inference, instead, we rely on GPT prompting with in-context learning examples to extract suitable descriptions. We report the prompt below, along the associated in-context examples in Figure~\ref{fig:icl}.\looseness=-1
\vspace{5pt}
\begin{tcolorbox}[colback=gray!5!white, colframe=gray!75!black, title=\textbf{Inference-time $c_\text{infer}$ generation prompt}, breakable, boxrule=0.5mm, colbacktitle=gray!75!black, coltitle=white]
\scriptsize{Reference:
The first five attached examples are the first frames of simulated videos that contain multiple objects on a surface, in each example one or two objects can start moving, and these objects can or cannot collide with other objects depending on their initial velocity and direction. The range of velocities for the initial motion of these objects is from 0.5m/s to 1.5m/s.\\
 
- In the first initial frame the green cube moves at 1.23m/s and the red cylinder at 1.07m/s. The entire video description for this video is: "On a wooden floor, there are four objects: a red cylinder, a green cube, a yellow cube, and a white cylinder. The red cylinder starts in the center and moves to the right, while the green cube moves to the left. The red cylinder continues to move to the right and eventually collides slightly with the white cylinder. The green cube continues to move left and collides slightly with the yellow cube. The red cylinder and the white cylinder remain stationary after the collision.". \\
 
- In the second initial frame the white cube moves at 0.67m/s and the purple cube at 1.35m/s, the entire description for this video is: "On a grassy background, there are a yellow cylinder, a purple cube, a pink cube, and a white cube. The white cube moves towards the yellow cylinder and collides slightly with it. The white cube then stops next to the yellow cylinder. The purple cube moves towards the pink cube and collides with it, moving it towards the top right, after the collision all objects remain stationary.".\\
 
- In the third initial frame the white cube moves at 1.5m/s and the light blue cube at 1.43m/s, the entire description for this video is: "Four cubes of different colors (yellow, green, blue, and purple) are on a textured, brown surface. The blue cube moves towards the green cube, colliding with it and pushing it downwards. The yellow cube moves towards the pink cube, they collide and the purple cube moves out of the frame. After the collision the blue and green cylinder are static next to each other, while the yellow cube is positioned in the top right part of the frame.".\\
 
- In the fourth initial frame, the gray cube moves at 0.64m/s, the entire description for this video is: "The scene is set on a brick ground with patches of green moss. A white cylinder is stationary in the background. A gray cube, initially positioned on the left side of the frame, moves horizontally to the right and eventually stops in the center of the frame. Two pink cubes, one on the right side and another slightly behind it, remain stationary throughout the sequence.".\\
 
- In the fifth initial frame the white cylinder moves at 0.71m/s and the pink cylinder moves at 0.55m/s. The entire description for this video is: "The scene consists of a grassy background with four 3D shapes: a purple cube, a white cylinder, a pink cylinder, and another purple cube. The white cylinder and the pink cylinder move slightly to the left. The purple cubes in the foreground remain stationary throughout the sequence.".\\
 
Objective:
-The last attached initial frame consists of a real picture, in this case the xxx on the xxx moves to the xxx at xxxm/s !!. According to this information and taking into account the provided examples from simulated videos, please provide an entire description, predicting what is going to happen in this scene. The provided video description should follow the same style and words used in the given simulated examples !! , reason about the real scene and describe if collisions happen or not, if they do, describe them, consider the distance and velocity between objects in the real scene to determine how the objects move, and how strong the collisions are. Make sure the description would include realistic effects if they happen produced by the movement or collisions that may occur between objects, for example movement of liquids, steam effects, falling objects, etc. If these effects are not present in the scene please do not include anything related to them in the prompt !!.  Do not include velocity values in it, friction effects, information from the surface, inclinations, rotations, or produced sounds.
}
\end{tcolorbox}

\vspace{-10pt}\subsubsection{Training Hyperparameters.}
We train \methodname using bf16 mixed precision for 1000 steps with a total batch size of 40 and saved checkpoints every 250 steps. For CogvideoX, our backbone is initialized from the pretrained \emph{THUDM/CogVideoX-5b-I2V} model, for Wan with \emph{Wan-AI/Wan2.2-TI2V-5B} and for Cosmos with \emph{nvidia/Cosmos-Predict2.5-2B} from their respective HuggingFace repositories. For the Controlnet we used the first 8 transformer layers, a downscaling factor of 8 and a control-weight of 0.5. We use AdamW optimizer with a learning rate of $1 \times 10^{-4}$, $\beta_1$=0.9, $\beta_2$=0.95, a cosine-with-restart learning-rate schedule and a max gradient norm of 1.0. \looseness=-1

\subsubsection{Baseline implementation details.} 
We compare our main \emph{KineMask-Cogvideo} model with Force Prompting, TORA, Motion-I2V, and the original models CogvideoX and Wan. All methods use an input prompt and an initial image. Additionally, Force Prompting takes the point coordinates of where to apply the force, the force prompt, and the angle of movement. Tora and Motion-I2V need the entire motion trajectory, we give the initial and final points. In all cases, we apply the same prompt $c_\text{infer}$ that we extract by querying GPT. In the extracted prompts, the moving object and its interactions are always described. \looseness=-1

\subsubsection{Prompts for Figure~\ref{fig:3d-impact}.}
We report here the prompts for rendering Figure~\ref{fig:3d-impact}, that we omitted due to the lack of space. \looseness=-1

\begin{tcolorbox}[colback=gray!5!white, colframe=gray!75!black, title=\textbf{Full Prompt Figure~\ref{fig:3d-impact} (left)}, breakable, boxrule=0.5mm, colbacktitle=gray!75!black, coltitle=white]
\scriptsize{"A ceramic pot moves left, falls from the chair to the floor, and completely shatters into many small pieces. After breaking, the pot disappears and only broken ceramic fragments remain scattered on the floor."}

\end{tcolorbox}

\begin{tcolorbox}[colback=gray!5!white, colframe=gray!75!black, title=\textbf{Full Prompt Figure~\ref{fig:3d-impact} (right)}, breakable, boxrule=0.5mm, colbacktitle=gray!75!black, coltitle=white]
\scriptsize{
"Two yellow rubber ducks are floating on a wet surface with shallow water. The duck on the left begins moving to the right, heading towards the duck on the right. As the duck on the left advances, small ripples spread outward from its base, disturbing the water. The duck on the left continues its motion and collides firmly with the duck on the right. The impact creates overlapping ripples and small splashes around both ducks. The duck on the right shifts to the side, while the duck on the left comes to a stop next to it. After the collision, the water ripples gradually spread and fade, and both ducks remain stationary."}
\end{tcolorbox}

\vspace{-10pt}\begin{figure*}[ht]
\centering
\resizebox{0.75\linewidth}{!}{
\includegraphics[width=0.3\linewidth]{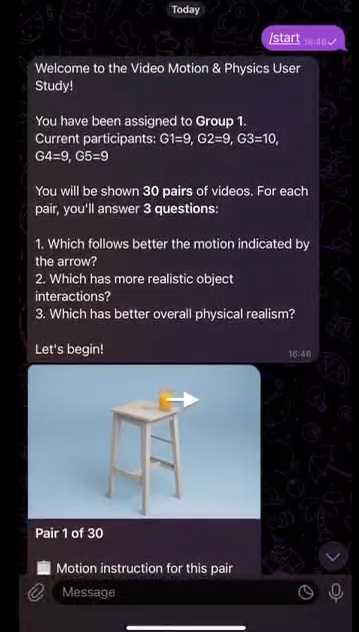}
\hfill
\includegraphics[width=0.3\linewidth]{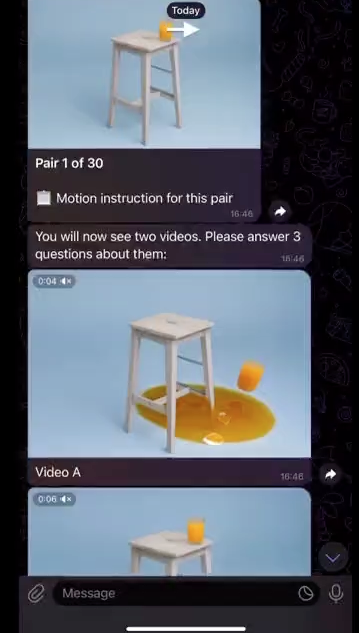}
\hfill
\includegraphics[width=0.3\linewidth]{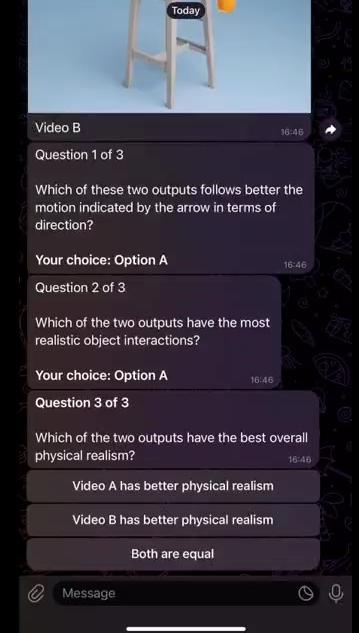}
}
\caption{\textbf{User study interface.} We collect pairwise user preference on direction, interactions realism, and consistency of objects.}\label{fig:telegram}
\vspace{-20pt}
\end{figure*}

\subsubsection{Details for user study.}
We now provide additional details from our user study. First, we divide users into 5 groups, randomizing the videos to display to limit the time necessary to complete our study. We display the videos in couples, along the original frame with the represented control, via a Telegram bot specifically used for the task and represented in Figure~\ref{fig:telegram}. For each video pair, we ask:\looseness=-1
\begin{enumerate}[topsep=0pt, itemsep=0pt, parsep=0pt, partopsep=0pt]
    \item Which of these two outputs follow better the motion indicated by the arrow in terms of direction?
    \item Which of the two outputs has the most realistic object interactions?
    \item Which of the two outputs have the best overall physical realism?
\end{enumerate}
We aggregate replies from all users in order to build the plots in Figure~\ref{fig:userstudy}.

\vspace{-10pt}
\subsection{Additional results}\label{sec:appendix-ablations}

\begin{figure*}[!htbp]
    \centering
    \includegraphics[width=0.95\linewidth]{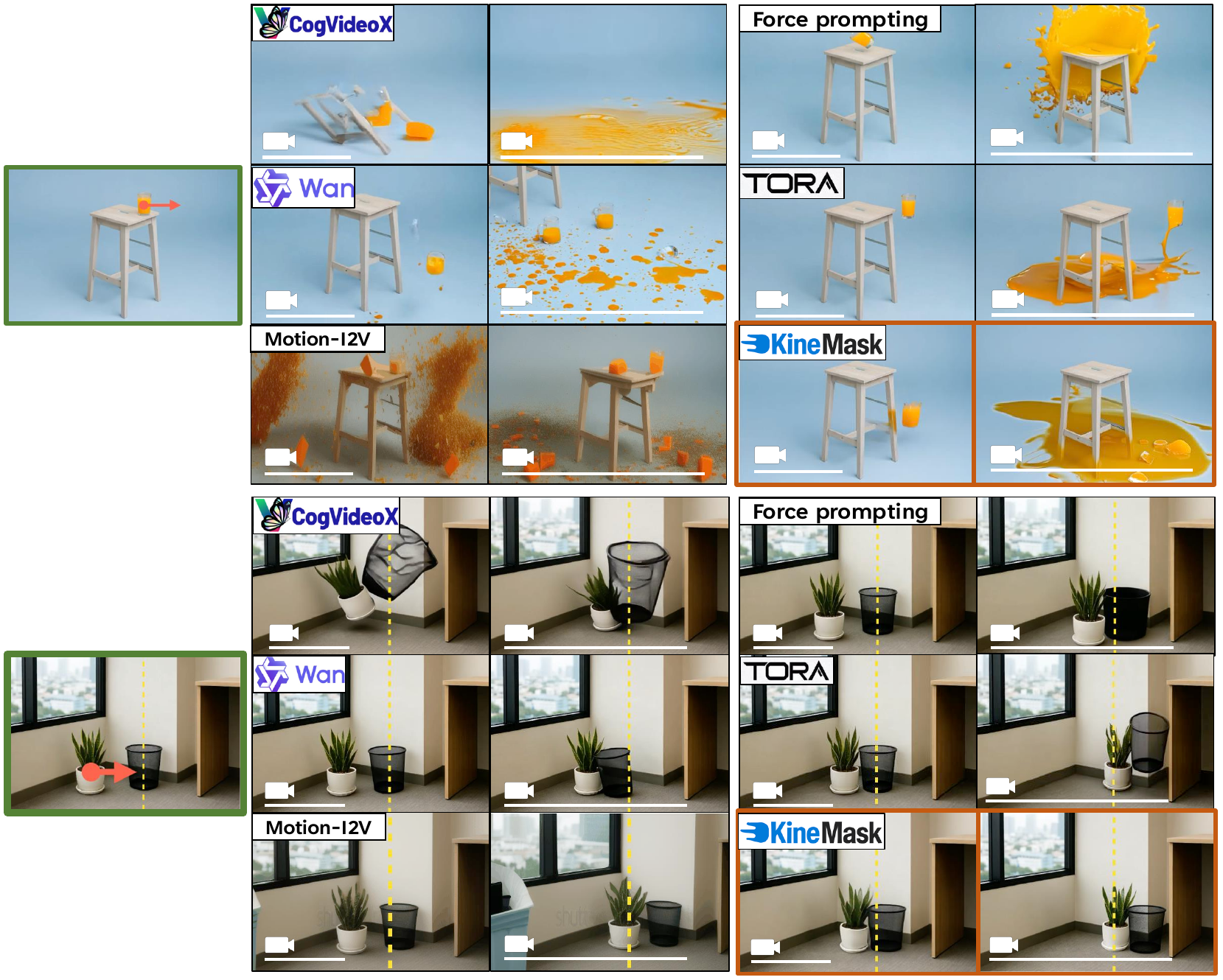}
    \vspace{0.1em}
    \includegraphics[width=0.95\linewidth]
    {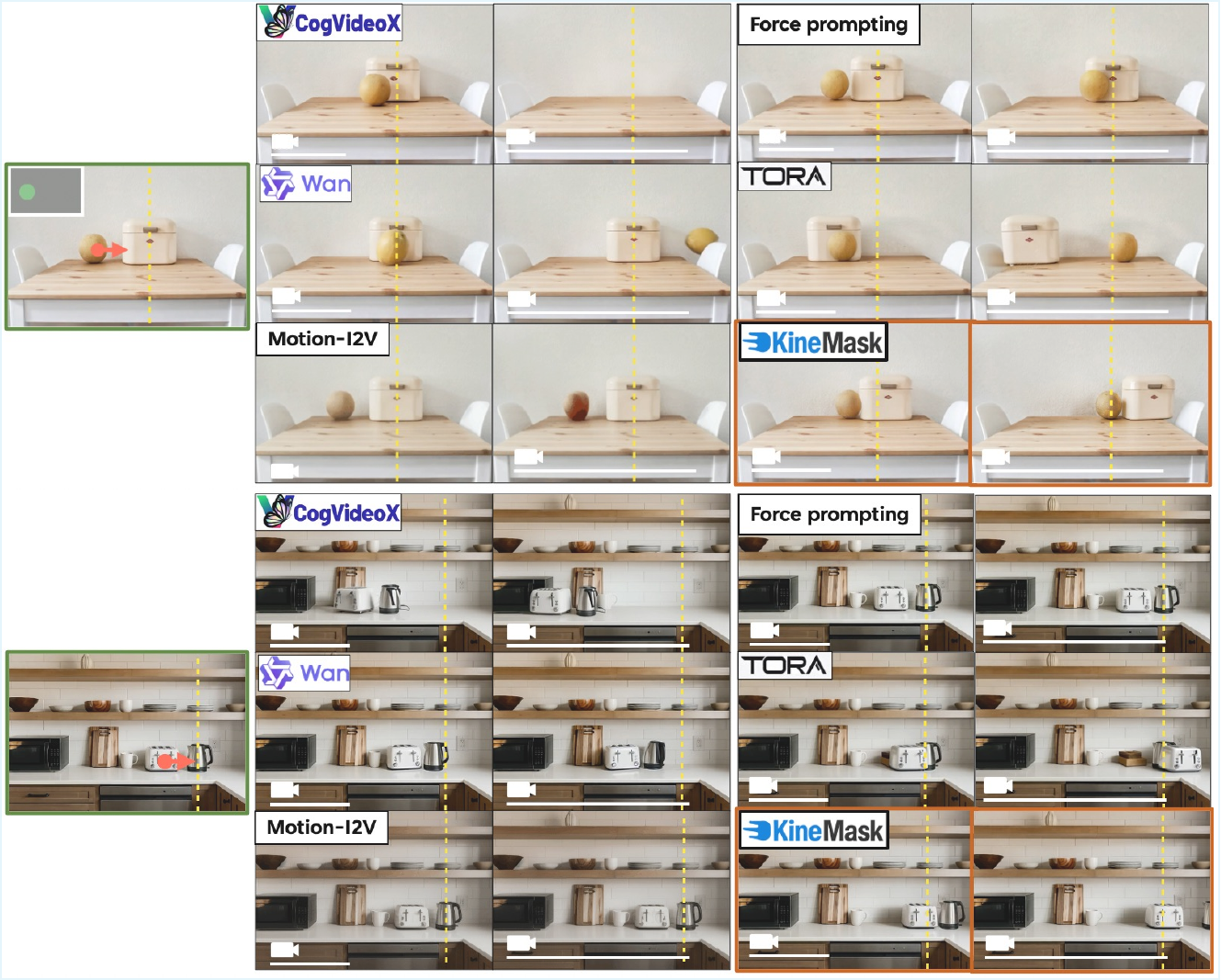}
    \caption{Qualitative comparison with all baselines.}
    \label{fig:Extraexamples}
\end{figure*}

\subsubsection{Qualitative comparison against all baselines.}
Figure~\ref{fig:Extraexamples} presents a complete qualitative comparison between \emph{KineMask-CogvideoX} and all baselines. Beyond the issues observed with CogVideoX (discussed in Section~\ref{sec:baselines}), Wan exhibits strong hallucinations and unrealistic object interactions. On the other hand, Drag-based methods such as Force-Prompting and Motion-I2V commonly suffer from slow object motion when objects are moved toward others, failing to synthesize proper interactions. Finally, TORA often produces hallucinations and ignores collisions. KineMask successfully generate plausible motion and object collisions according to the user input, showing its capability to correctly understand causality, and dynamics created due to collisions, and showing the generalization of the knowledge obtained from synthetic data in complex real-world scenes, even when some 3D understanding is necessary.\looseness=-1

\vspace{-8pt}\subsubsection{Additional Qualitative Comparison.}
\begin{figure*}[t]
    \centering
    \begin{minipage}[t]{0.49\linewidth}
        \centering
        \includegraphics[width=\linewidth]{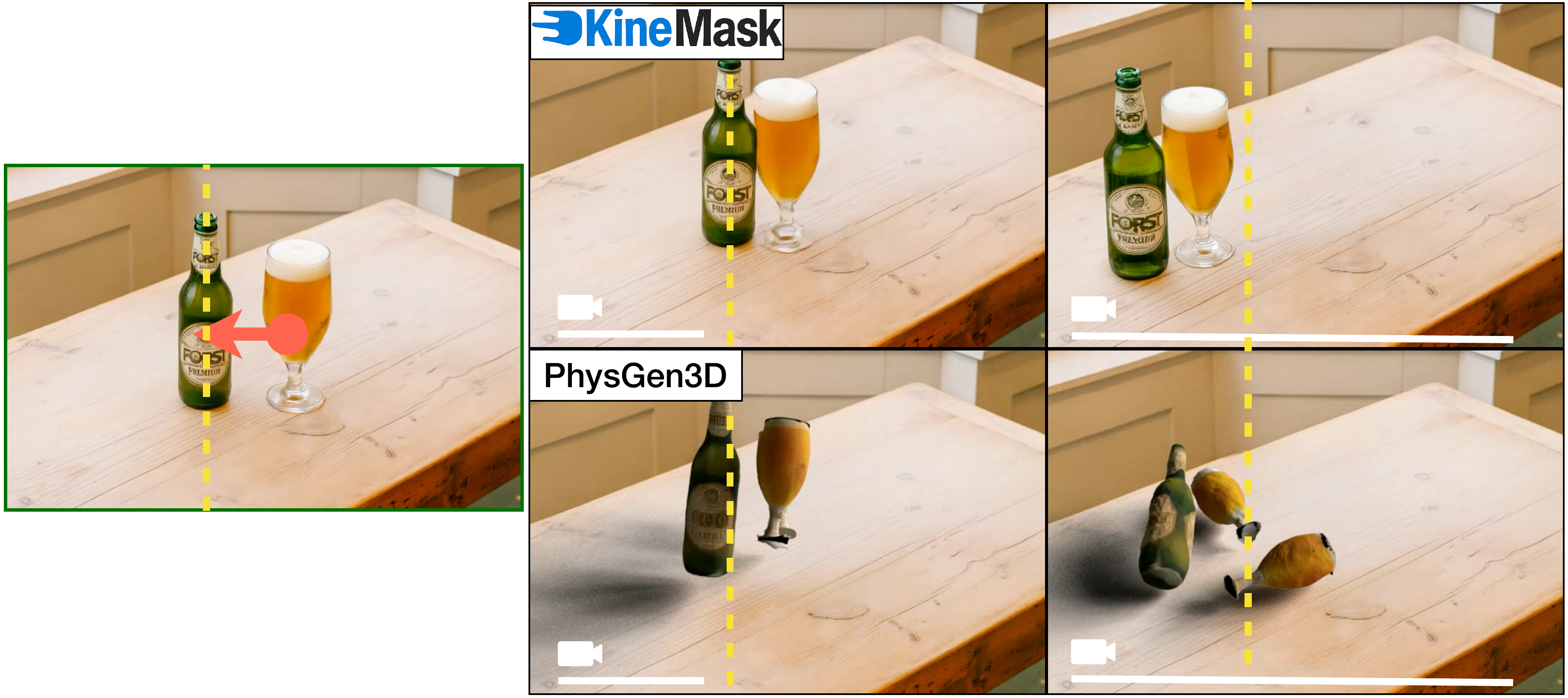}
    \end{minipage}\hfill
    \begin{minipage}[t]{0.49\linewidth}
        \centering
        \includegraphics[width=\linewidth]{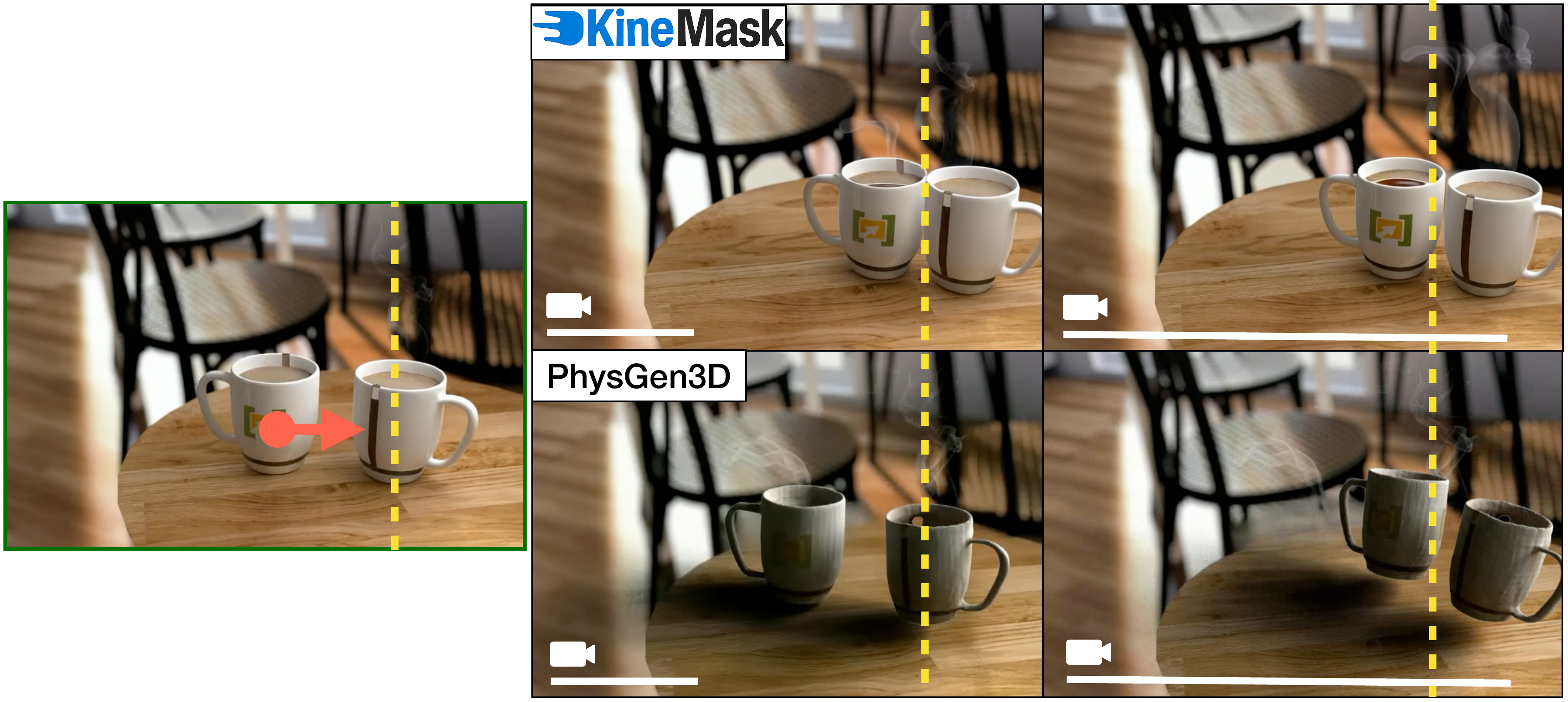}
    \end{minipage}

    \caption{Qualitative comparison of KineMask vs PhysGen3D.}
    \label{fig:phys_comp}
    \vspace{-5pt}
\end{figure*}

In addition to comparing with our initial set of baselines, we also perform an additional qualitative comparison with the recent PhysGen3D ~\cite{chen2025physgen3d}, which uses 3D information, a simulator in the loop and a diffusion model as a rendering tool to generate plausible videos from an initial image. While this brings plausible physics due to the dynamics simulator, this causes multiple problems and limitations, from the complexity of the pipeline, dependency on the 3D recontruction, and the lack of visual realism. As shown in Figure~\ref{fig:phys_comp}, PhysGen3D can be vulnerable and dependent on its 3D reconstruction setup, which can cause incorrect object reconstructions, such as creating duplicated objects, causing a chain of failures towards synthesizing proper rigid-body interactions. Beyond the synthetic appearance of objects, PhysGen3D and simulation-based methods also are unreliable to generate effects like liquid or steam moving as a result of a collision, as shown in the second example where two cups collide. On the other hand, KineMask synthesizes plausible object collisions and effects created from object interactions, showing its understanding of rigid-body dynamics.\looseness=-1

\begin{table}[!t]
    \setlength{\tabcolsep}{4pt}
    \centering
    \resizebox{0.6\linewidth}{!}{%
    \begin{tabular}{ll|cccc}
        \multicolumn{2}{c}{} & \multicolumn{4}{c}{\interactions} \\
        \toprule
        \textbf{Backbone} & \textbf{Setup} &
        \textbf{MSE}$\downarrow$ & \textbf{FVD}$\downarrow$ &
        \textbf{FMVD}$\downarrow$ & \textbf{IoU}$\uparrow$ \\
        \midrule
        \multirow{2}{*}{Wan} & Baseline & 431.4 & 977.7 & 3649.8 & 0.192 \\
        & KineMask & \textbf{211.0} & \textbf{475.6} & \textbf{968.4} & \textbf{0.257} \\
        \midrule
        \multirow{2}{*}{CogVideoX} & Baseline & 344.6 & 807.3 & 3514.9 & 0.192 \\
        & KineMask & \textbf{158.7} & \textbf{250.7} & \textbf{143.8} & \textbf{0.355} \\
        \midrule
        \multirow{2}{*}{Cosmos} & Baseline & 1463.6 & 1287.3 & 1285.7 & 0.203 \\
        & KineMask & \textbf{445.1} & \textbf{433.6} & \textbf{385.6} & \textbf{0.306} \\
        \bottomrule
    \end{tabular}%
    }
    \vspace{5pt}
    \caption{\textbf{Comparison of different model backbones.} We compare all versions of KineMask using different base models.}
    \label{tab:cogvswan}
    \vspace{-20pt}
\end{table}

\vspace{-8pt}\subsubsection{Test on different model backbones}
As discussed in Section~\ref{sec:analysis}, we tested the generalization of our proposed approach on two additional model backbones Wan2.2-5B and Cosmos2.5-2B. Table~\ref{tab:cogvswan} shows some quantitative results on the test set of our synthetic data, comparing all versions of KineMask against the original model backbones, it can be seen that all  KineMasks versions outperform their original backbones by large margins respectively. \looseness=-1

\vspace{-8pt}\subsubsection{Additional Quantitative Comparison.}
\begin{figure*}[!t]
    \centering
    \includegraphics[width=\linewidth]{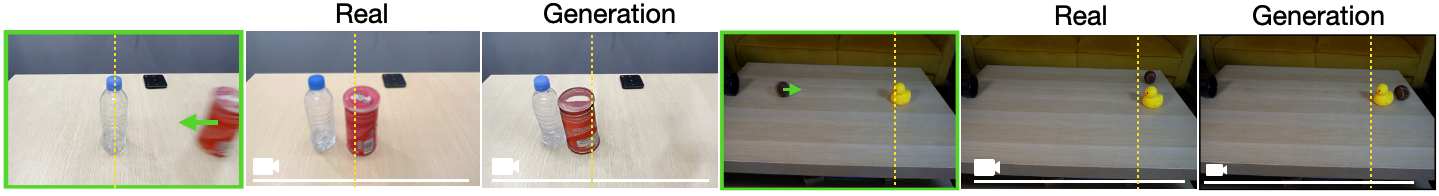}
    \vspace{-15pt}
    \caption{\textbf{Additional Comparisons on Real-World Videos.} Real and Generation examples for our own data (left) and Physics-IQ (right)} \vspace{-12pt}
    \label{fig:comprealvideos}
    \vspace{10pt}
\end{figure*}

Our human evaluation in Section~\ref{sec:baselines} focused on realistic scenes with no available ground truth; therefore, in order to provide a quantitative comparison in real-world scenes with rigid body interactions, we use real-world videos from PhysicsIQ~\cite{motamed2025generative} and videos recorded in our lab. In Table~\ref{tab:addquantieval} we compare KineMask-CogVideoX with the main baselines and an additional finetuned CogvideoX I2V on synthetic data. KineMask still outperforms all, coherently with our user evaluation. Real and generation examples for our own data (left) and Physics-IQ (right) are shown in Figure\ref{fig:comprealvideos}.\looseness=-1

\begin{table}[!t]
\centering
\setlength{\tabcolsep}{4pt}
\resizebox{\linewidth}{!}{%
\begin{tabular}{l|cccc|cccc}
  \toprule
  & \multicolumn{4}{c|}{\textbf{Physics IQ}} & \multicolumn{4}{c}{\textbf{Our Data}} \\
  \textbf{Method} & \textbf{MSE}$\downarrow$ & \textbf{FVD}$\downarrow$ & \textbf{FMVD}$\downarrow$ & \textbf{IoU}$\uparrow$ & \textbf{MSE}$\downarrow$ & \textbf{FVD}$\downarrow$ & \textbf{FMVD}$\downarrow$ & \textbf{IoU}$\uparrow$ \\
  \midrule
  Force-Prompting & \textbf{27.76} & 652.5 &860.4  &0.358 &8029.4 &469.7 & 1833.9 & 0.217 \\
  TORA & 28.05  & 667.0 & 1327.6 & 0.330 & 9273.1 & 465.2 & 4107.7 & 0.193 \\
  CogvideoX (zero-shot) & 88.87 & 581.3 & 1608.1 & 0.282 & 11367.5 & 609.4 & 3404.4 & 0.227 \\
  CogvideoX (finetuned) & 65.86 & 357.1 &\textbf{ 331.6} & 0.220 & 6032.2 & 319.22 & \textbf{405.62} &  0.201 \\
  KineMask & {39.67} & {\bfseries 212.1} & \underline{431.81} & {\bfseries 0.428} & \textbf{5551.9} & \textbf{295.7}& \underline{422.38} & \textbf{0.339} \\
  \bottomrule
\end{tabular}%
}
\vspace{3pt}
\caption{\textbf{Additional Quantitative Results.} We report additional comparisons for our main model Kinemask-CogVideoX against the main baselines and the zero-shot and finetuned backbone model CogvideoX.}
\label{tab:addquantieval}
\vspace{-20pt}
\end{table}

\vspace{-8pt}\subsubsection{Velocity encoding.}
We propose an ablation on our encoding mechanism. In \methodname, we assume to encode the instantaneous velocity of objects in frame $t$, hence $v_t$. This leads to barely visible masks in the presence of low velocity magnitude when the velocity decreases in presence of interactions. We have tested another configuration, by encoding instead for each frame the initial velocity $v_0$, thus having well-defined masks for all the motion trajectory. However, from the results in Table~\ref{tab:ablation-a}, we see that encoding the per frame velocity $v_t$ yields better results. We speculate that this strategy would help the model understand the decrease trend of velocity that naturally occurs in motion, ultimately serving as an implicit regularization term for training.\looseness=-1

\begin{table}[t]
    \centering

    \begin{subtable}[t]{0.44\linewidth}
        \centering
        \setlength{\tabcolsep}{2pt}
        \renewcommand{\arraystretch}{1.2}
        \resizebox{\linewidth}{!}{
        \begin{tabular}{l|cccc}
            \multicolumn{1}{c}{}& \multicolumn{4}{c}{\textit{Interactions}} \\
            \toprule
            \textbf{Encoding} & \textbf{MSE}$\downarrow$ & \textbf{FVD}$\downarrow$ & \textbf{FMVD}$\downarrow$ & \textbf{IoU}$\uparrow$ \\
            \midrule
            $v_0$ & 205.7 & 277.3 & 216.3 & 0.294 \\
            $v_t$ (ours) & \textbf{160.9} & \textbf{231.3} & \textbf{174.4} & \textbf{0.376} \\
            \bottomrule
        \end{tabular}}
        \subcaption{Velocity encoding}
        \label{tab:ablation-a}
    \end{subtable}
    \hfill
    \begin{subtable}[t]{0.49\linewidth}
        \centering
        \setlength{\tabcolsep}{2pt}
        \renewcommand{\arraystretch}{1.2}
        \resizebox{\linewidth}{!}{
        \begin{tabular}{l|cccc}
            \multicolumn{1}{c}{}& \multicolumn{4}{c}{\textit{Interactions}} \\
            \toprule
            \textbf{Sampling} & \textbf{MSE}$\downarrow$ & \textbf{FVD}$\downarrow$ & \textbf{FMVD}$\downarrow$ & \textbf{IoU}$\uparrow$ \\
            \midrule
            Uniform & 232.0 & 303.1 & 190.9 & 0.323 \\
            Non-uniform (ours) & \textbf{160.9} & \textbf{231.3} & \textbf{174.4} & \textbf{0.376} \\
            \bottomrule
        \end{tabular}}
        \subcaption{Non-uniform sampling}
        \label{tab:ablation-b}
    \end{subtable}

    \caption{\textbf{Additional ablation studies.} We report an additional encoding strategies for our conditioning masks in Table (\subref{tab:ablation-a}), proving that our design choice of using $v_t$ is best. Furthermore, we showcase that removing our non-uniform sampling on interactions-rich frames during dropout harms performance (\subref{tab:ablation-b}).}
    \label{tab:ablation-appendix}
    \vspace{-15pt}
\end{table}

\vspace{-8pt}\subsubsection{Dropout density.}
Section~\ref{sec:setup} mentioned that we sample non-uniformly the frame $f^*$ for our second-stage training. In Table~\ref{tab:ablation-b}, we present a quantitative evaluation of the effects of this design. As visible, sampling the dropout around the frames where collisions occur in the training set helps \methodname to focus on interaction-rich representations.\looseness=-1

\vspace{-8pt}\subsubsection{Role of Text}
\begin{figure*}[!t]
    \centering
    \includegraphics[width=\linewidth]{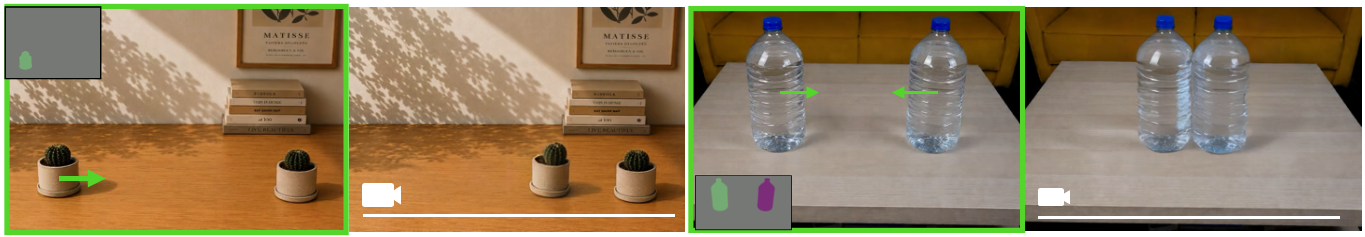}
    \vspace{-15pt}
    \caption{\textbf{Role of Text.} Generation examples when the mask and prompt disagree. Left, the prompt describes a collision while the velocity is not enough to create the interaction. Right, the prompt does not describe an interaction but the velocities of both bottles are enough to create a collision.}
    \label{fig:roleoftext}
    \vspace{-10pt}
\end{figure*}

We performed ablations when masks and prompts disagree. Figure~\ref{fig:roleoftext} left, shows no interaction when the velocity is not enough to create a collision between the two pots, but the prompt describes it \textit{"..The cactus pot on the left continues to move right and collides.."}. On the other hand, right shows interactions even though the prompt do not describe it, \textit{"On a table, there are two bottles."}. This shows that KineMask is not simply following VLM-triggered heuristics, furthermore, Figure~\ref{fig:causality} shows that changing the input velocity of the moving object changes the resulting dynamics. We noticed text mostly helps for visual realism and object consistency, but not as the sole source of behavior.\looseness=-1

\vspace{-8pt}\subsubsection{Qualitatives}
\begin{figure*}[t]
    \centering
    \resizebox{\textwidth}{!}{%
    \begin{minipage}{\textwidth}
        \centering
        \includegraphics[width=0.22\linewidth]{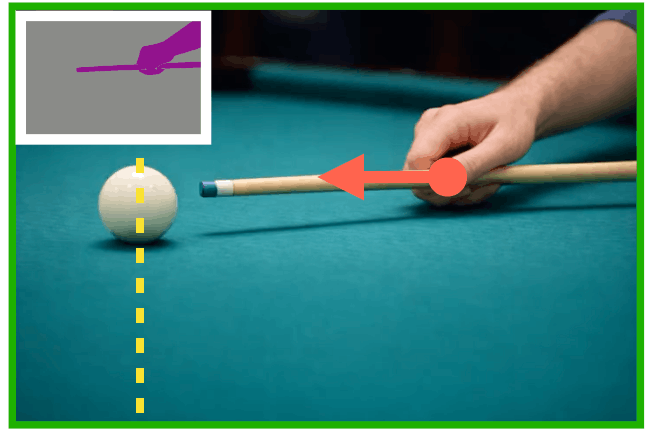}
        \includegraphics[width=0.22\linewidth]{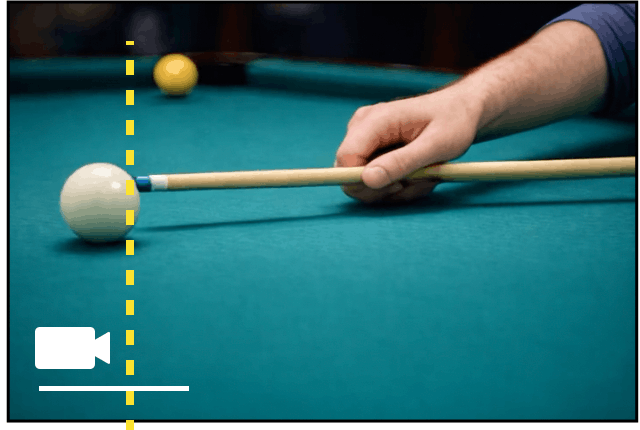}
        \includegraphics[width=0.22\linewidth]{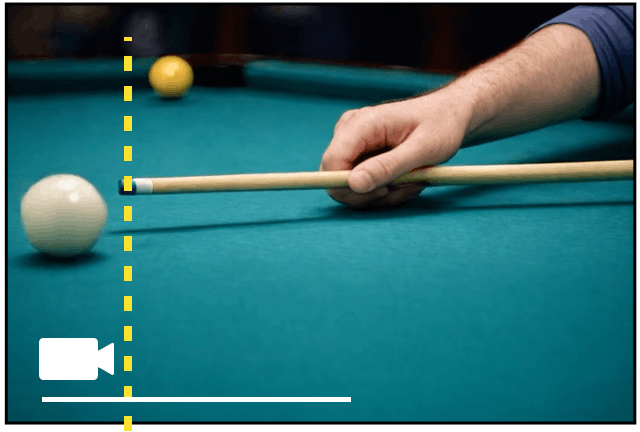}
        \includegraphics[width=0.22\linewidth]{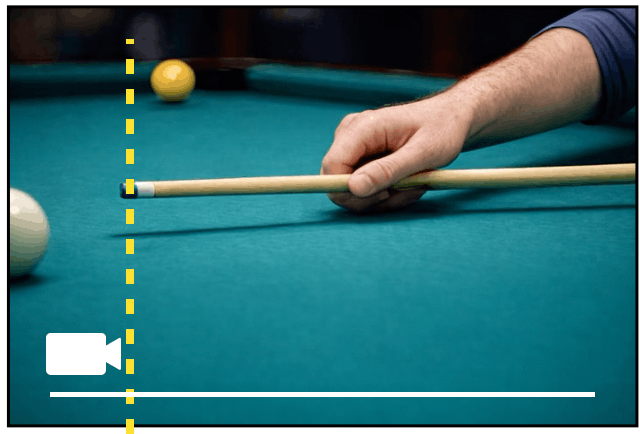}\\[2pt]
        
        \includegraphics[width=0.22\linewidth]{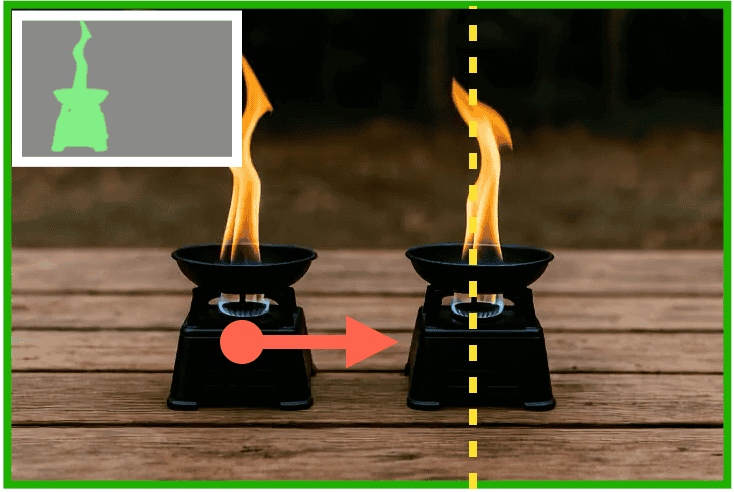}
        \includegraphics[width=0.22\linewidth]{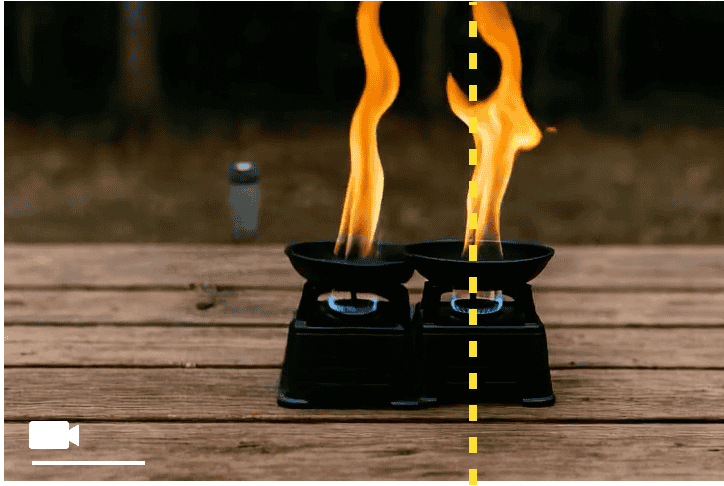}
        \includegraphics[width=0.22\linewidth]{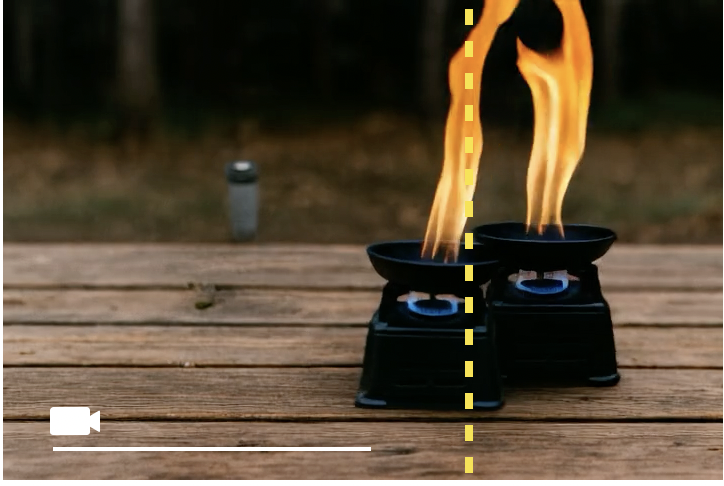}
        \includegraphics[width=0.22\linewidth]{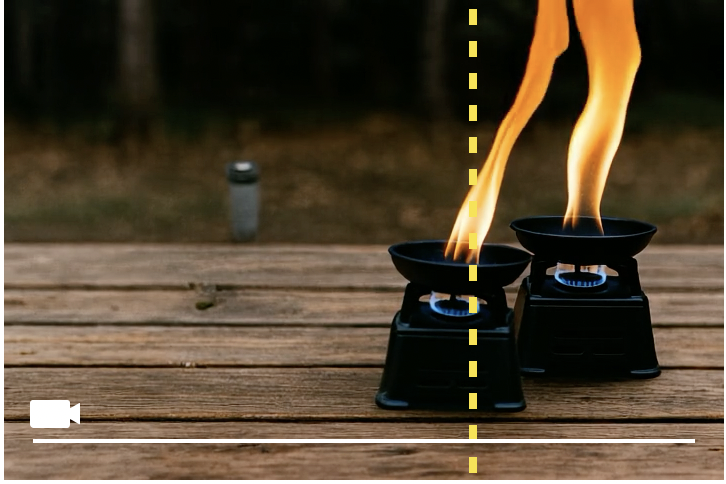}\\[2pt]

        \includegraphics[width=0.22\linewidth]{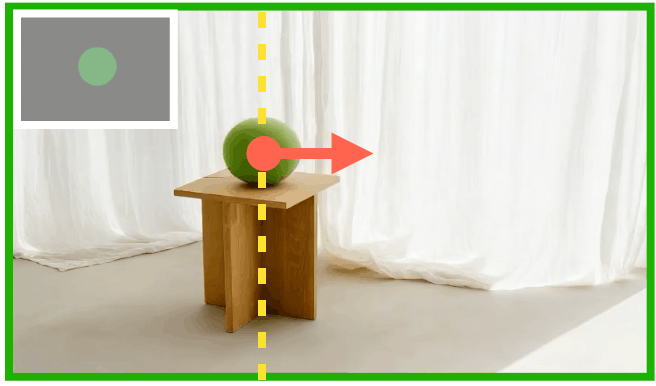}
        \includegraphics[width=0.22\linewidth]{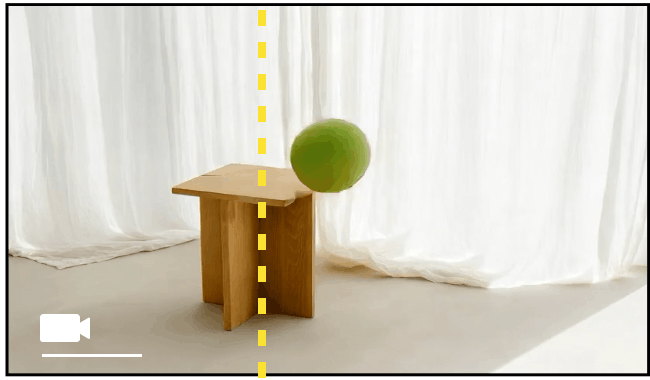}
        \includegraphics[width=0.22\linewidth]{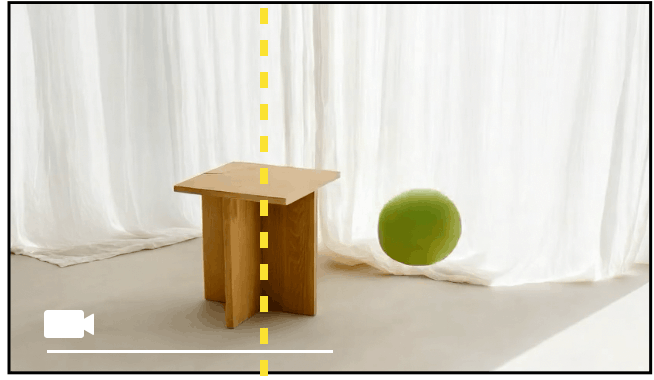}
        \includegraphics[width=0.22\linewidth]{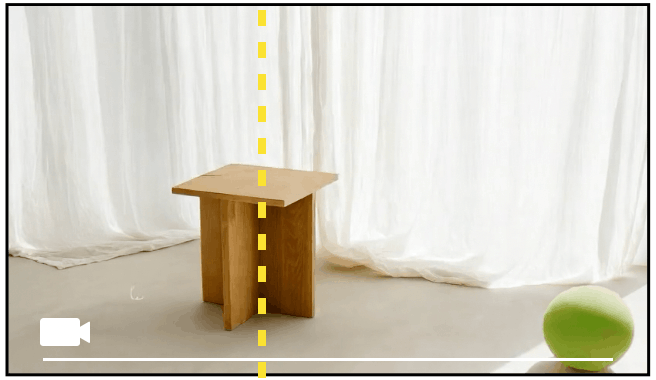}\\[2pt]

        \includegraphics[width=0.22\linewidth]{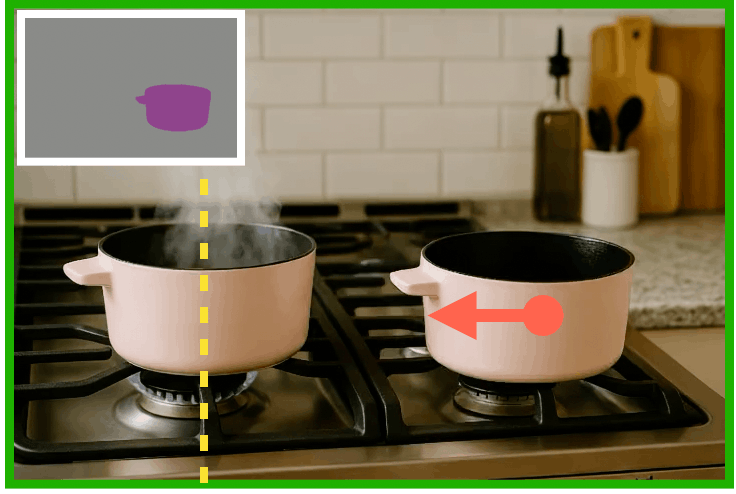}
        \includegraphics[width=0.22\linewidth]{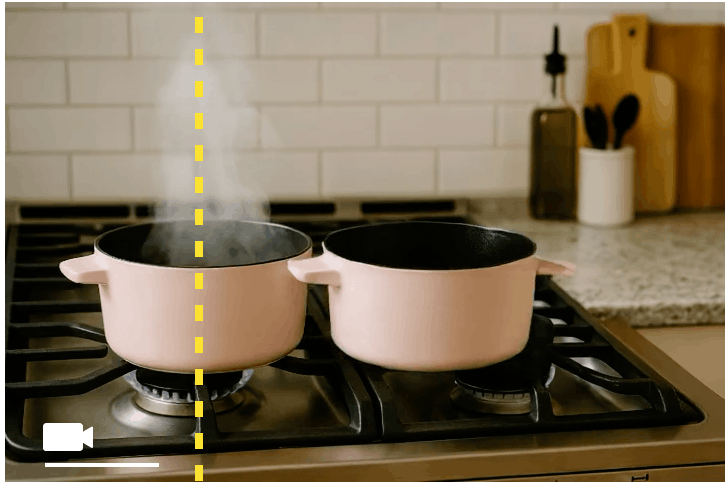}
        \includegraphics[width=0.22\linewidth]{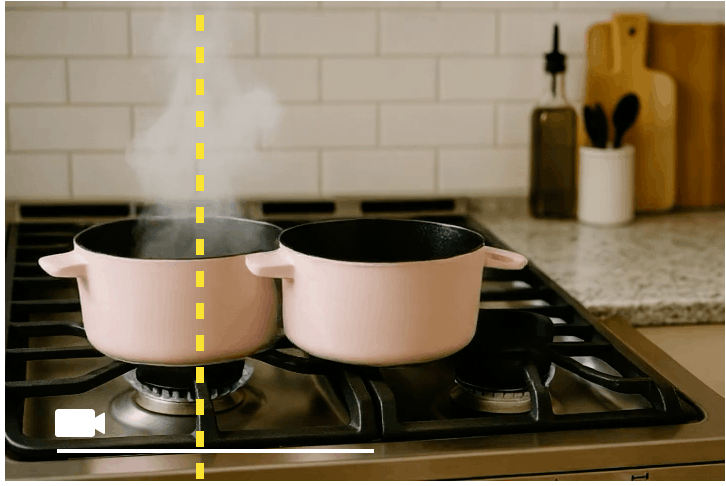}
        \includegraphics[width=0.22\linewidth]{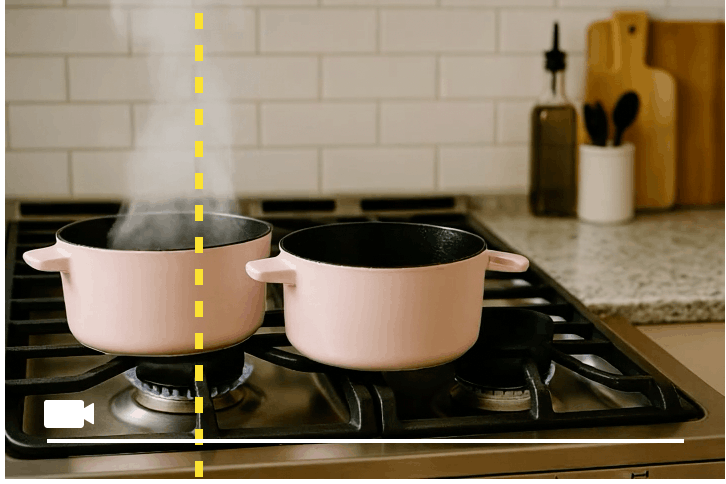}\\[2pt]
        
        \includegraphics[width=0.22\linewidth]{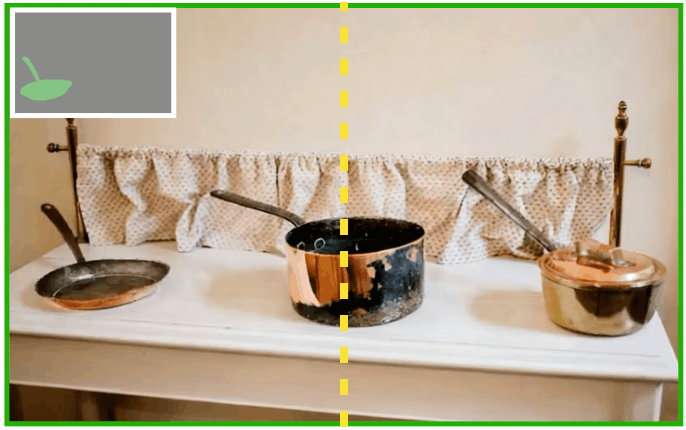}
        \includegraphics[width=0.22\linewidth]{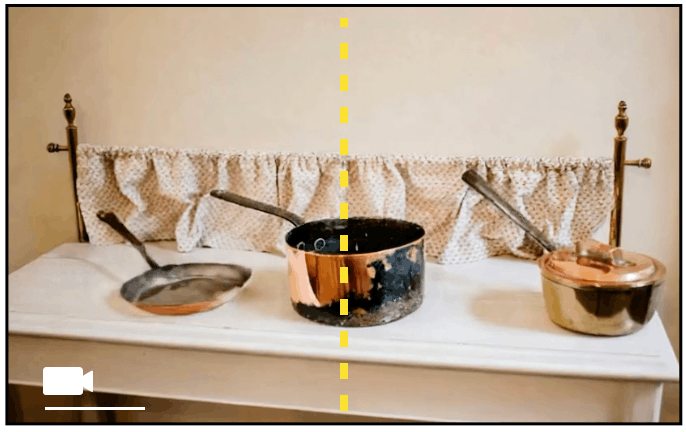}
        \includegraphics[width=0.22\linewidth]{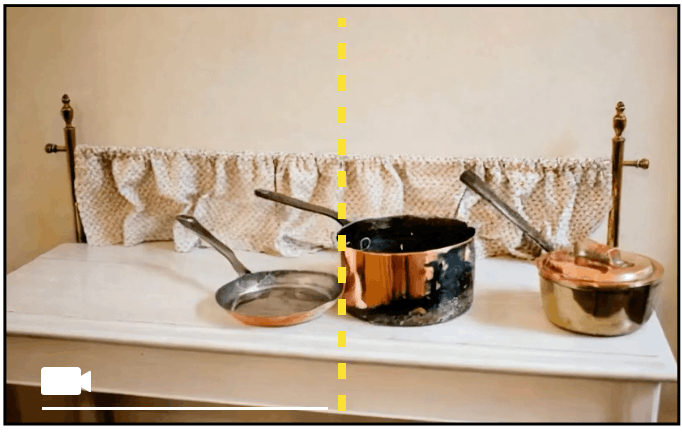}
        \includegraphics[width=0.22\linewidth]{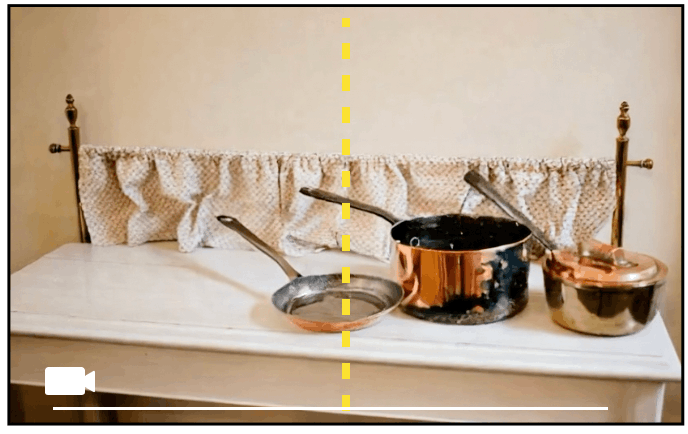}\\[2pt]


        \includegraphics[width=0.22\linewidth]{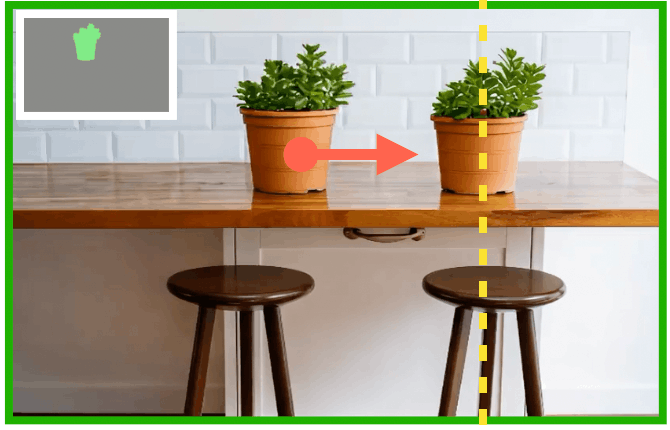}
        \includegraphics[width=0.22\linewidth]{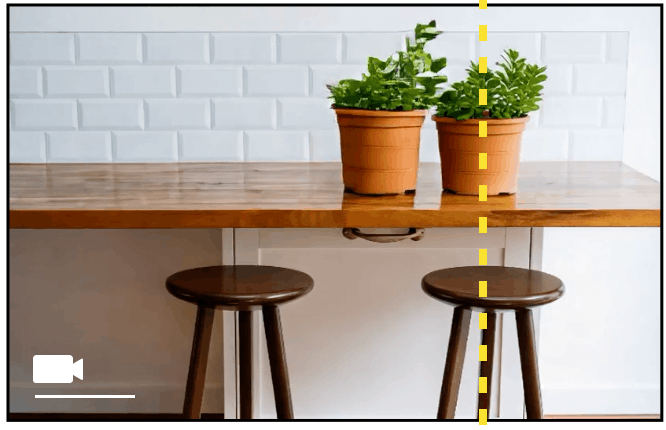}
        \includegraphics[width=0.22\linewidth]{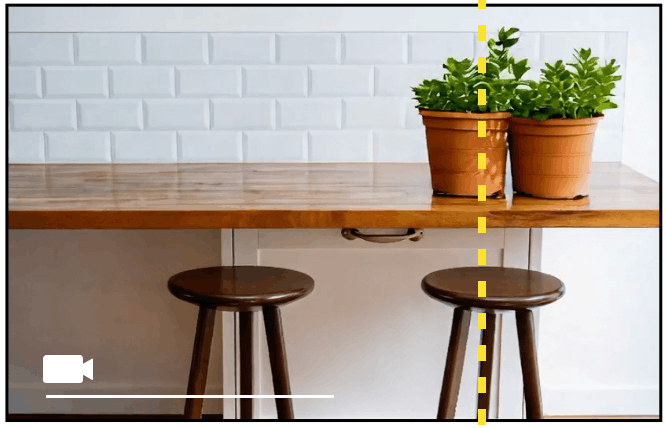}
        \includegraphics[width=0.22\linewidth]{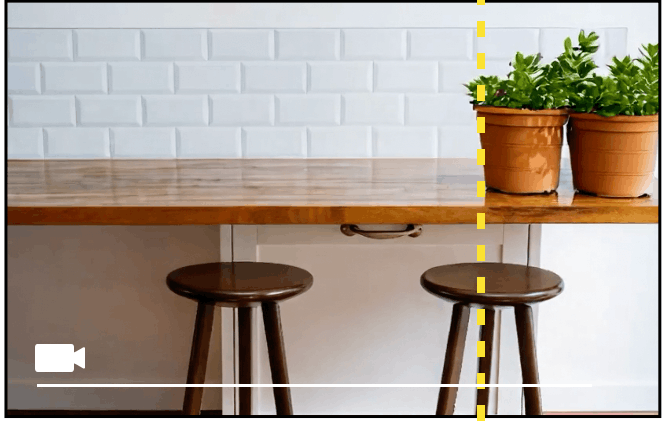}\\[2pt]

        \includegraphics[width=0.22\linewidth]{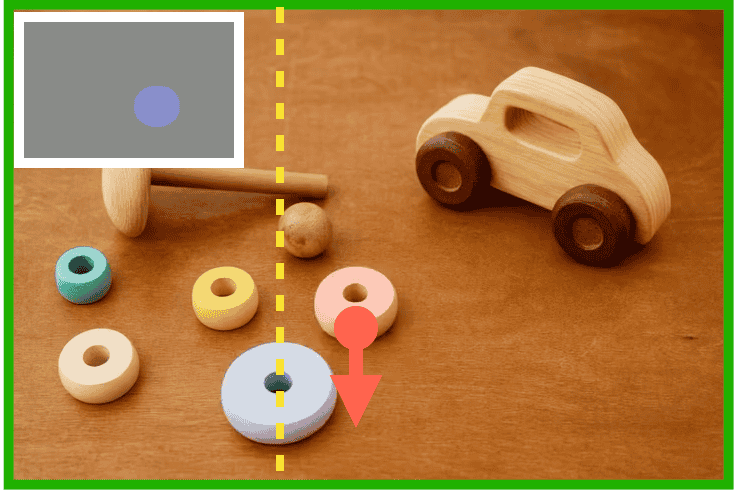}
        \includegraphics[width=0.22\linewidth]{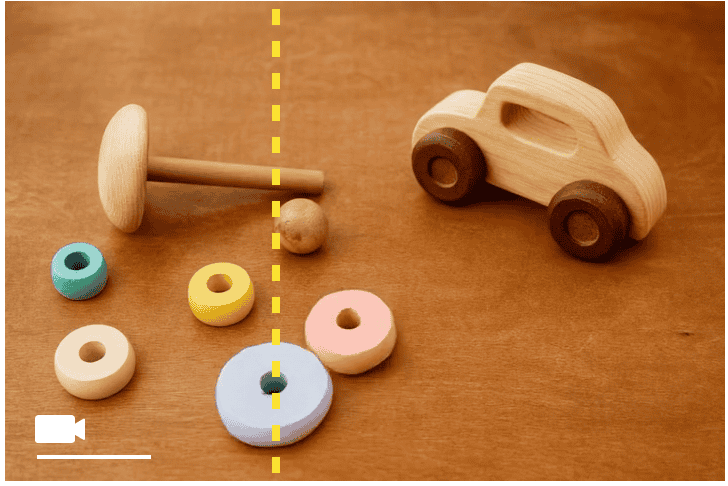}
        \includegraphics[width=0.22\linewidth]{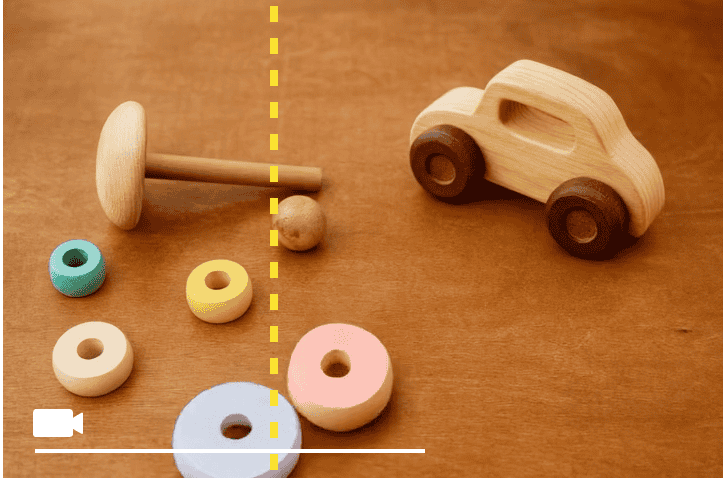}
        \includegraphics[width=0.22\linewidth]{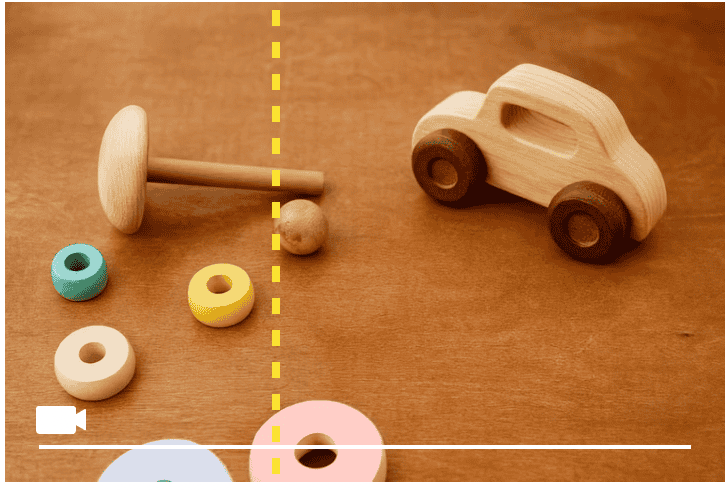}\\[2pt]



        \includegraphics[width=0.22\linewidth]{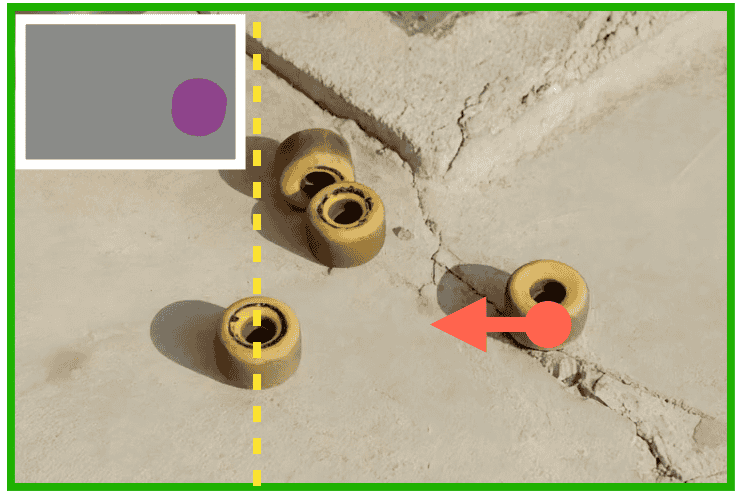}
        \includegraphics[width=0.22\linewidth]{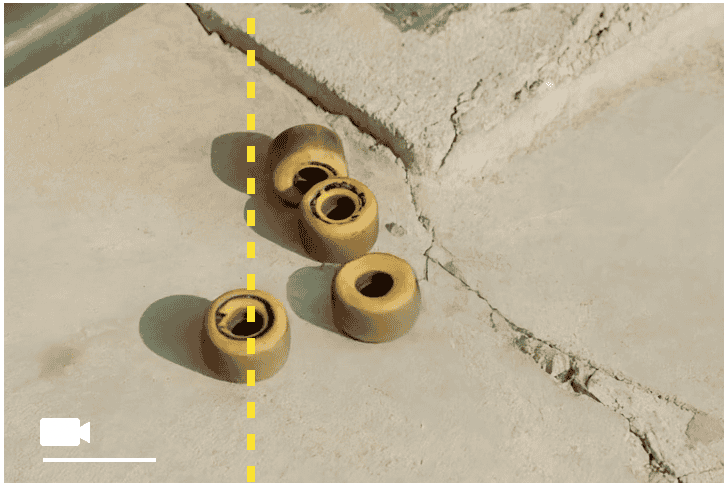}
        \includegraphics[width=0.22\linewidth]{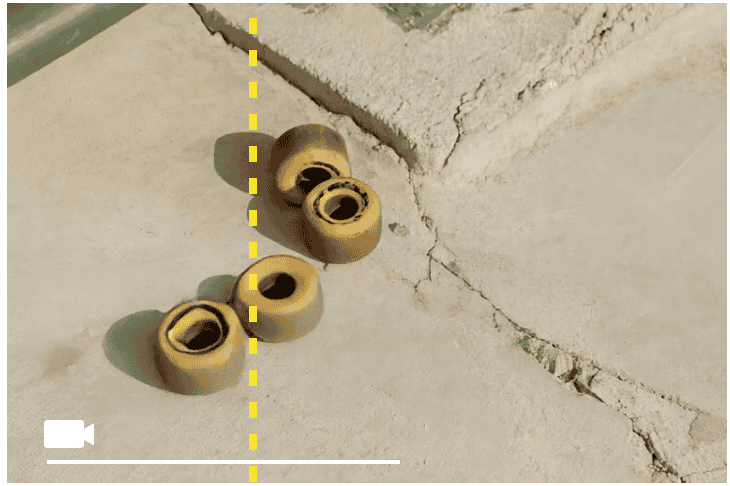}
        \includegraphics[width=0.22\linewidth]{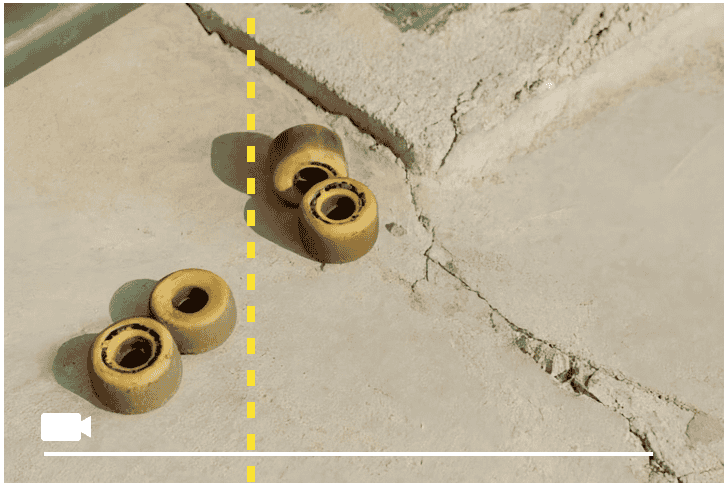}\\[2pt]

        \includegraphics[width=0.22\linewidth]{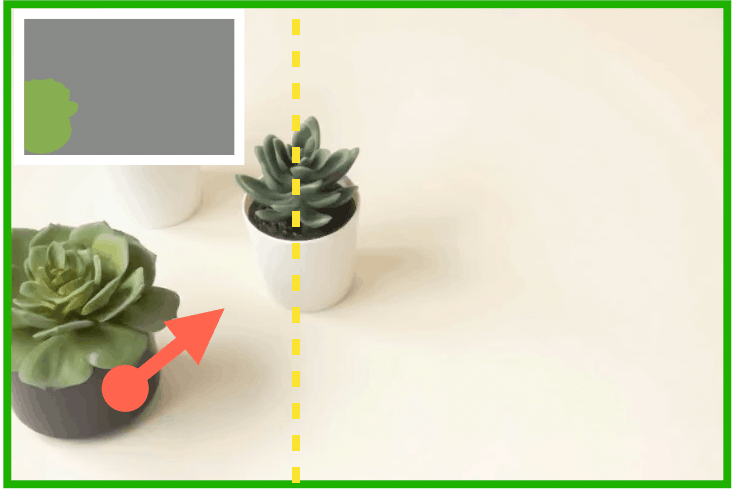}
        \includegraphics[width=0.22\linewidth]{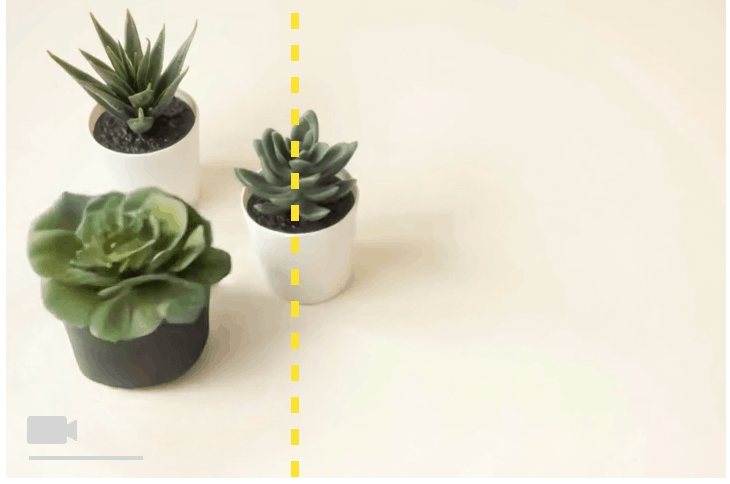}
        \includegraphics[width=0.22\linewidth]{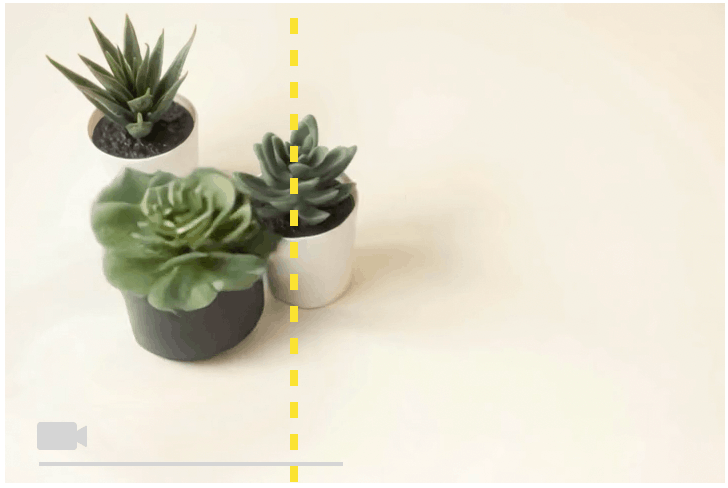}
        \includegraphics[width=0.22\linewidth]{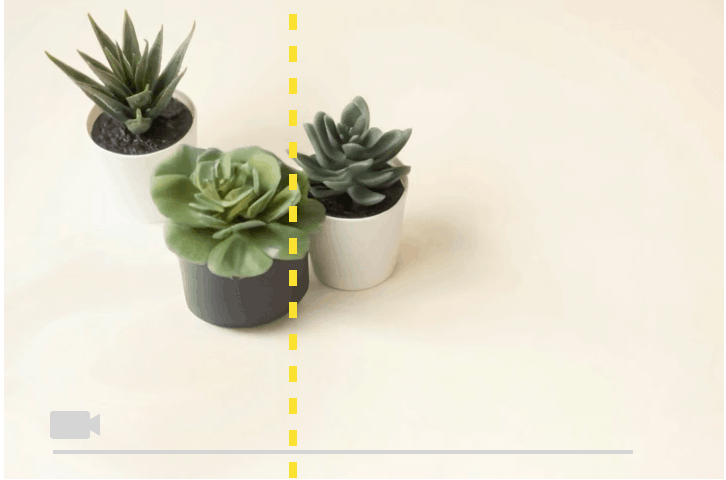}\\[2pt]
    \end{minipage}}
    \caption{Additional qualitative examples of \methodname.}
    \label{fig:Examples}
\end{figure*}

For ease of visualization, we report in Figure~\ref{fig:Examples} some additional qualitative results of interactions rendered with \methodname. As visible, in several real scenarios, we are able to render realistic object interactions, where objects move coherently with the users' prompt. \looseness=-1

\vspace{-8pt}\subsubsection{Handling complex objects}
\begin{figure*}[!t]
\centering
\setlength{\tabcolsep}{1pt}

\resizebox{1\linewidth}{!}{%
\begin{tabular}{@{}c@{\hspace{8pt}}c@{}}
\begin{tabular}{ccc}
\fcolorbox{darkgreen}{white}{\includegraphics[width=\linewidth]{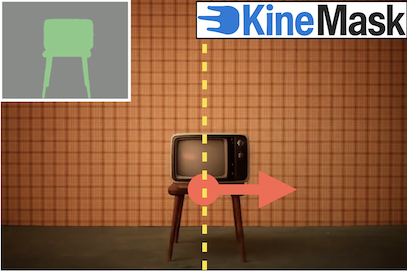}} &
\includegraphics[width=\linewidth]{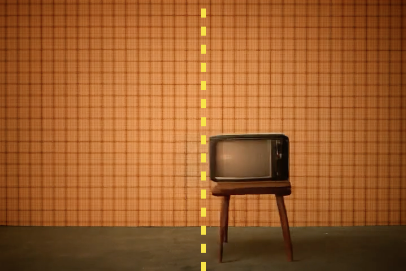} &
\includegraphics[width=\linewidth]{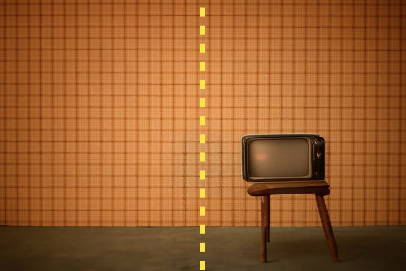} \\[-2pt]

\fcolorbox{darkgreen}{white}{\includegraphics[width=\linewidth]{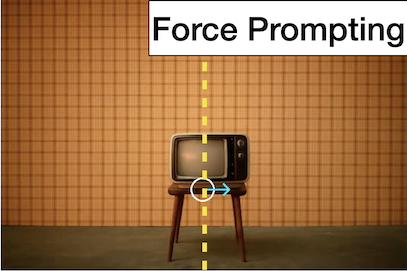}} &
\includegraphics[width=\linewidth]{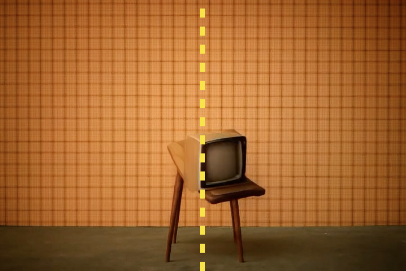} &
\includegraphics[width=\linewidth]{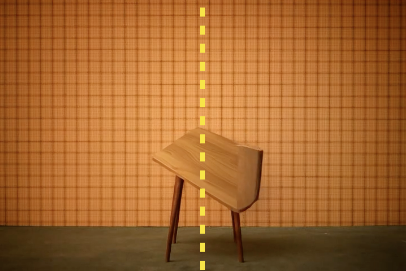} \\[-2pt]

\fcolorbox{darkgreen}{white}{\includegraphics[width=\linewidth]{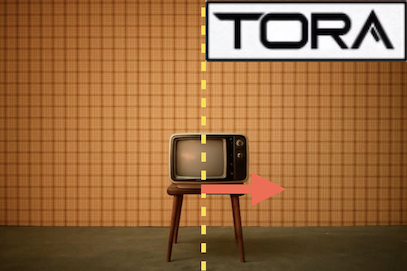}} &
\includegraphics[width=\linewidth]{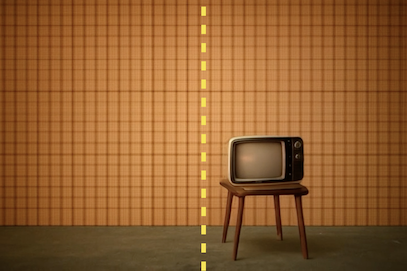} &
\includegraphics[width=\linewidth]{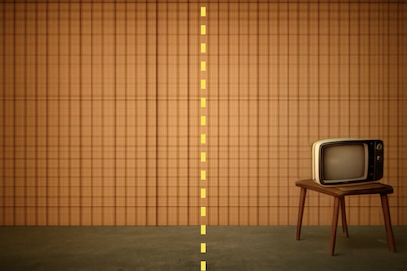} \\[-2pt]

\fcolorbox{darkgreen}{white}{\includegraphics[width=\linewidth]{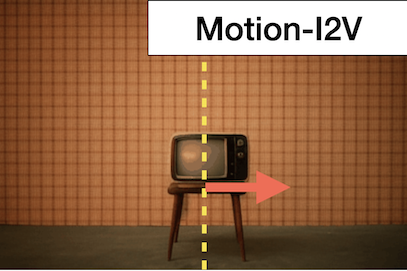}} &
\includegraphics[width=\linewidth]{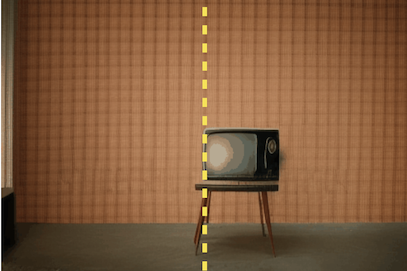} &
\includegraphics[width=\linewidth]{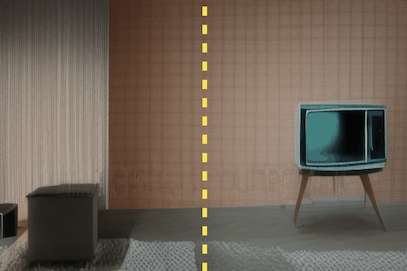} \\
\end{tabular}
&
\begin{tabular}{ccc}
\fcolorbox{darkgreen}{white}{\includegraphics[width=\linewidth]{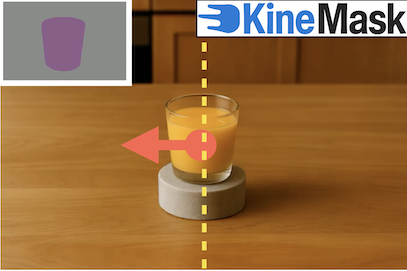}} &
\includegraphics[width=\linewidth]{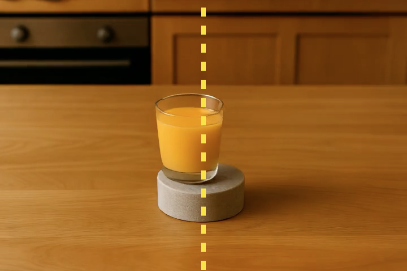} &
\includegraphics[width=\linewidth]{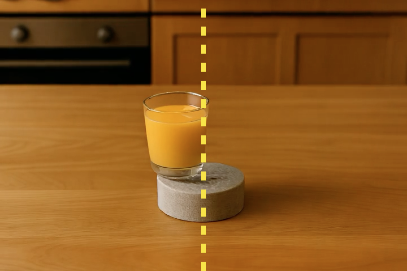} \\[-2pt]

\fcolorbox{darkgreen}{white}{\includegraphics[width=\linewidth]{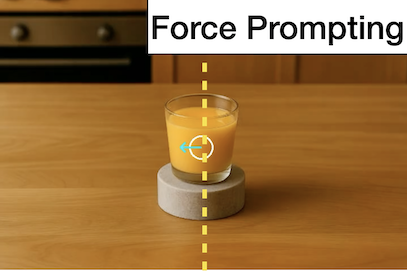}} &
\includegraphics[width=\linewidth]{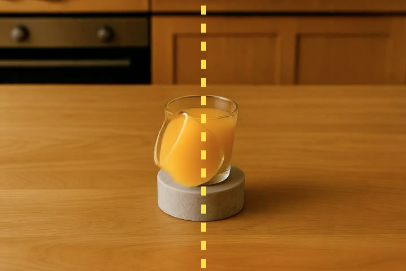} &
\includegraphics[width=\linewidth]{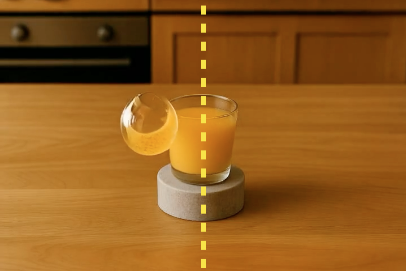} \\[-2pt]

\fcolorbox{darkgreen}{white}{\includegraphics[width=\linewidth]{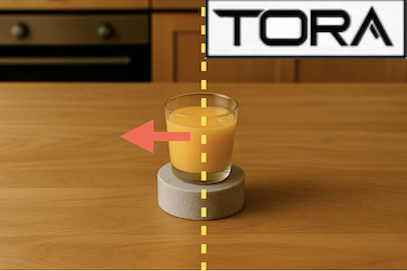}} &
\includegraphics[width=\linewidth]{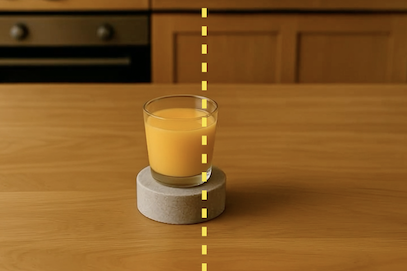} &
\includegraphics[width=\linewidth]{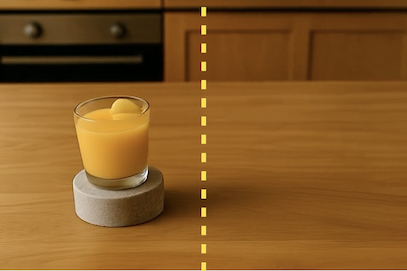} \\[-2pt]

\fcolorbox{darkgreen}{white}{\includegraphics[width=\linewidth]{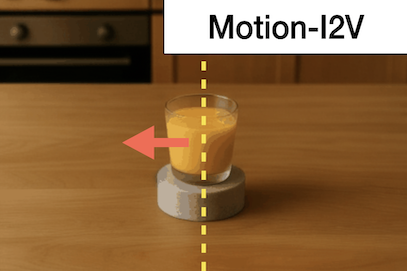}} &
\includegraphics[width=\linewidth]{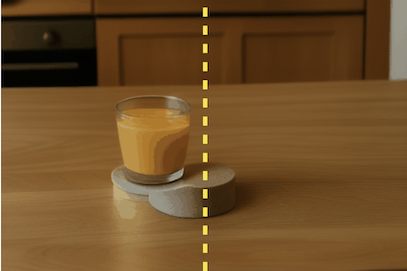} &
\includegraphics[width=\linewidth]{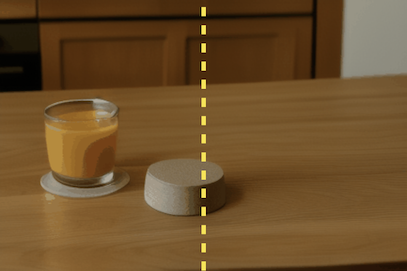} \\
\end{tabular}
\end{tabular}%
}

\caption{\textbf{Motion control.} Our mask-based control is robust to ambiguous scenes, while others suffer from hallucinations due to the ambiguous control signal.}
\label{fig:motion-control-comparison}
\vspace{-8pt}
\end{figure*}

In Figure~\ref{fig:motion-control-comparison}, we report an interesting failure mode of drag-based baselines succesfully handled by \methodname. As visible, Force Prompting, TORA and Motion-I2V struggle when the control is applied to object with thin support structures, or with piled objects. This is the result of an architectural decision, since their control mechanism does not employ masks for identifying the object to move with pixel level precision. Instead, \methodname correctly generates motion on these edge cases. This shows how the presence of a dense mask brings a lot of benefits to perform better motion control in ambiguous cases compared to sparse conditioning information.\looseness=-1

\vspace{-8pt}\subsubsection{Failure Cases}
\begin{figure*}[!t]
    \centering
    \scalebox{1}{
        \begin{minipage}{\textwidth}
        \centering
        \includegraphics[width=0.22\linewidth]{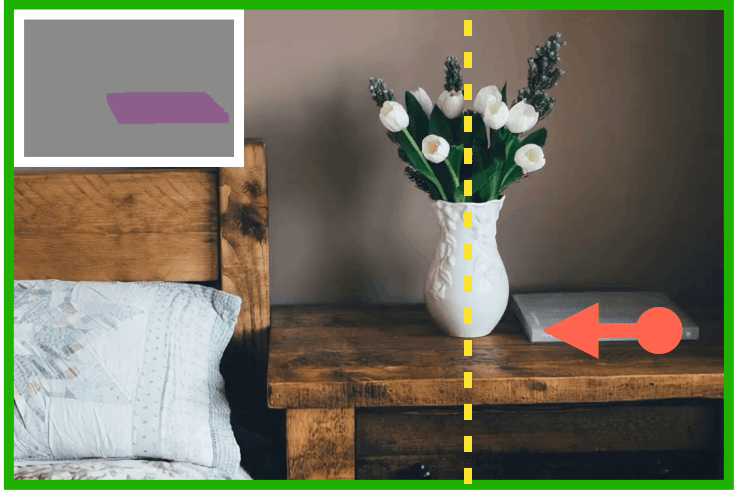}
        \includegraphics[width=0.22\linewidth]{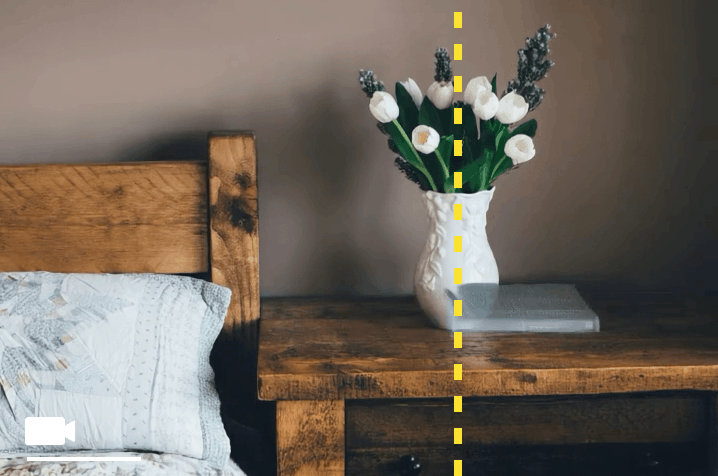}
        \includegraphics[width=0.22\linewidth]{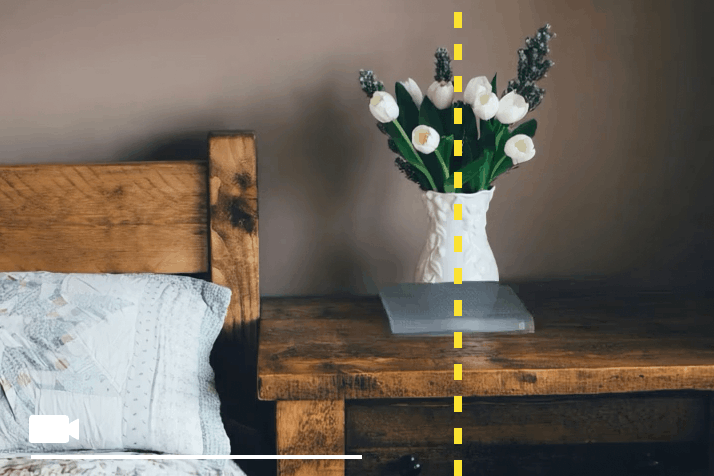}
        \includegraphics[width=0.22\linewidth]{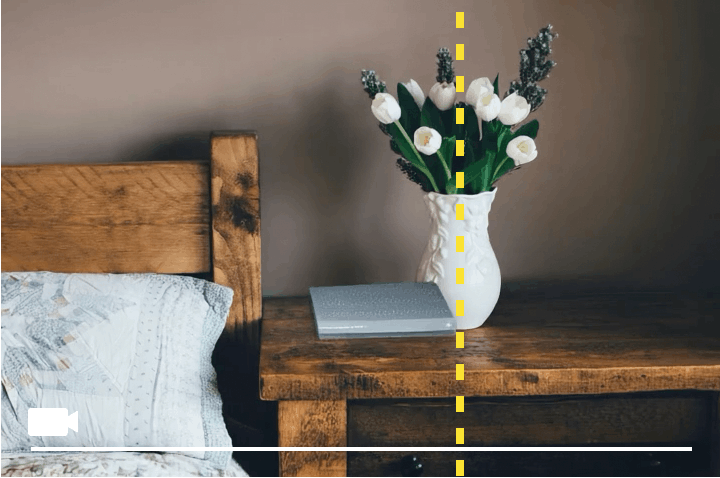}
        \\[2pt]
        \includegraphics[width=0.22\linewidth]{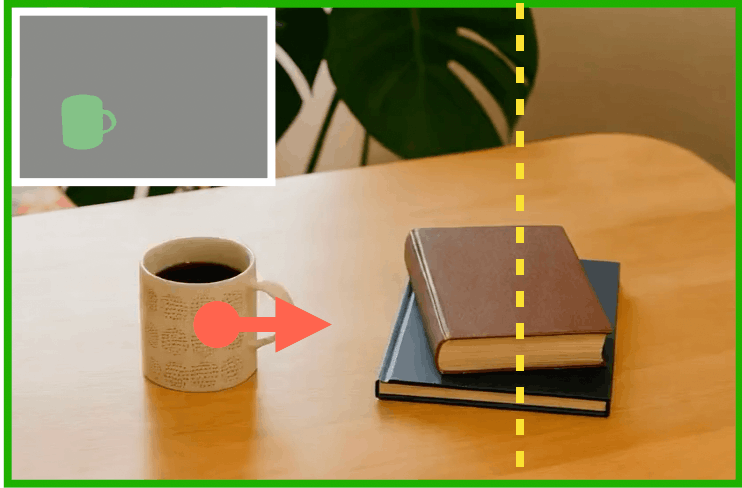}
        \includegraphics[width=0.22\linewidth]{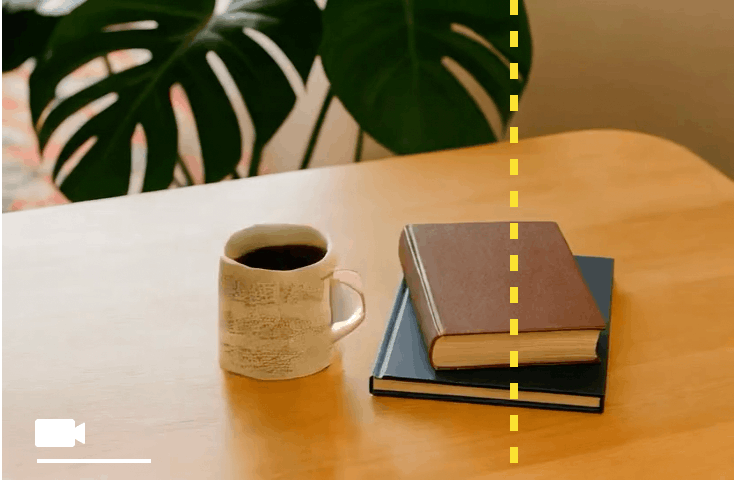}
        \includegraphics[width=0.22\linewidth]{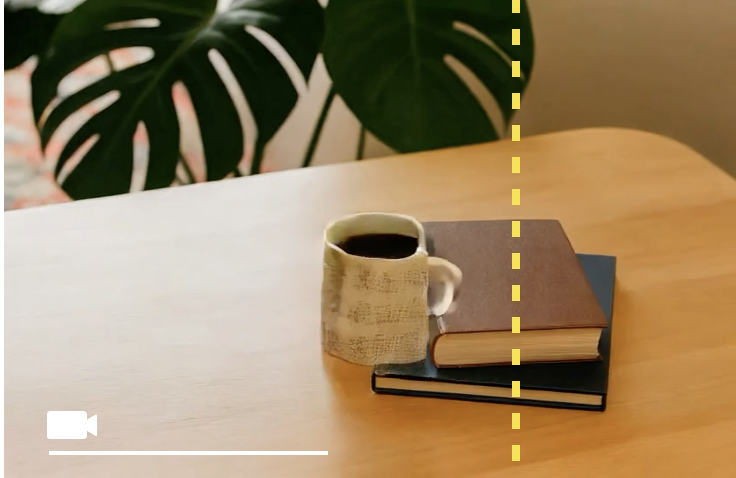}
        \includegraphics[width=0.22\linewidth]{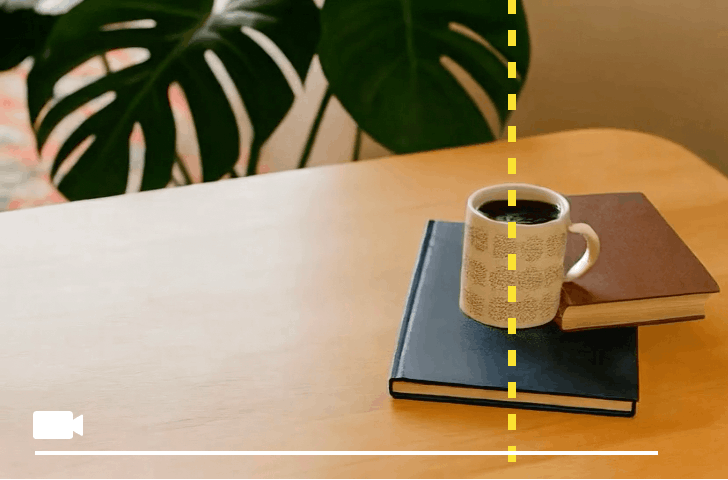}
        \\[2pt]
        \includegraphics[width=0.22\linewidth]{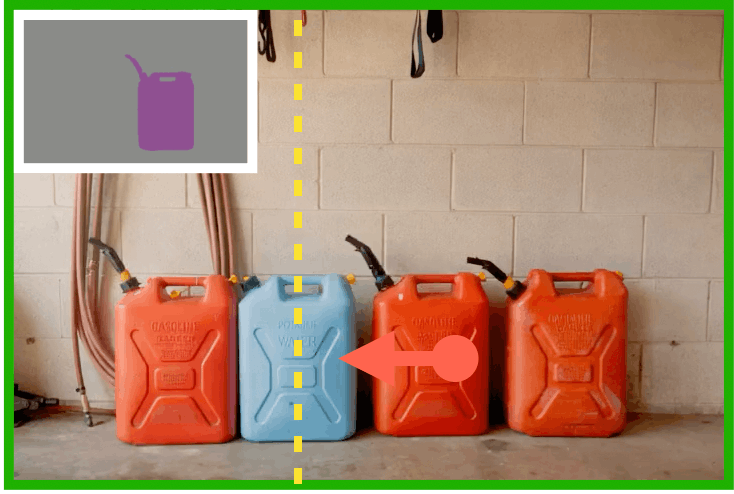}
        \includegraphics[width=0.22\linewidth]{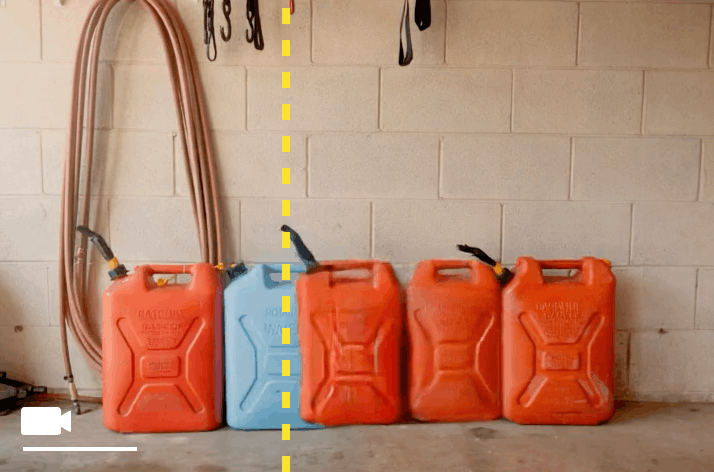}
        \includegraphics[width=0.22\linewidth]{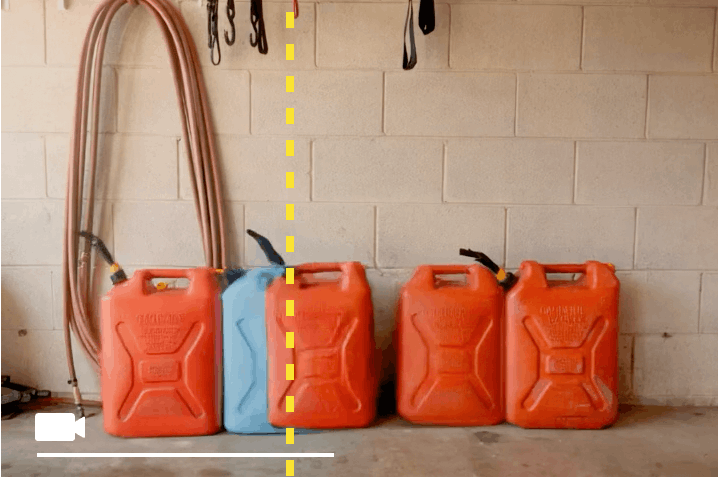}
        \includegraphics[width=0.22\linewidth]{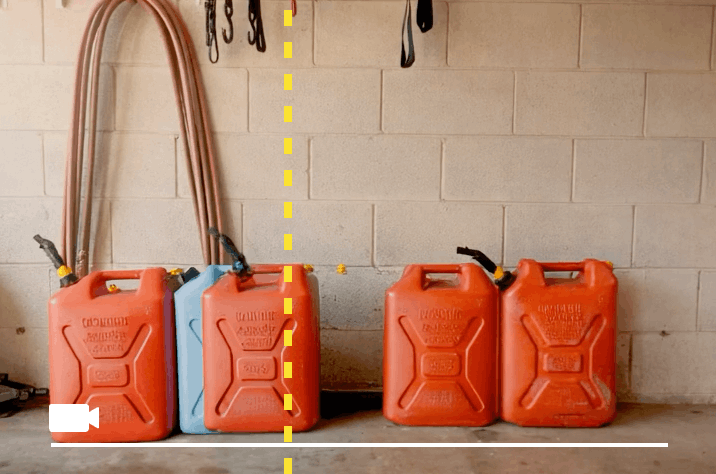}
        \\[2pt]
        \includegraphics[width=0.22\linewidth]{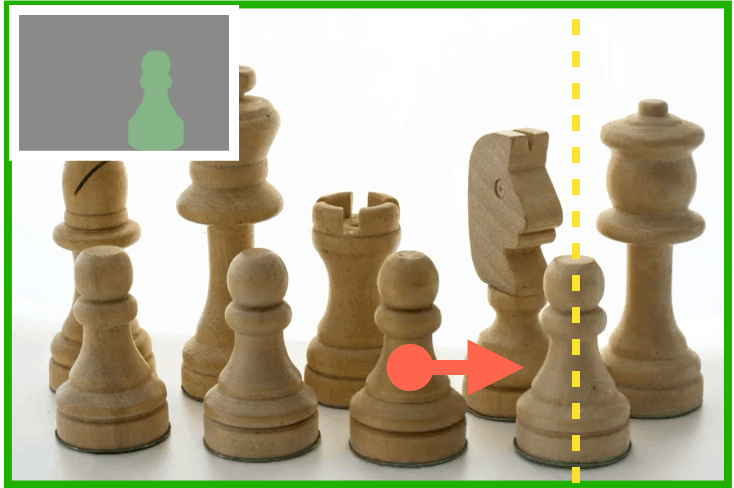}
        \includegraphics[width=0.22\linewidth]{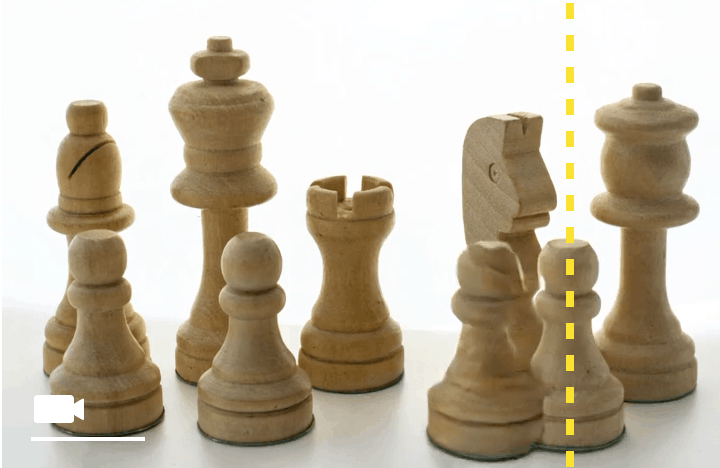}
        \includegraphics[width=0.22\linewidth]{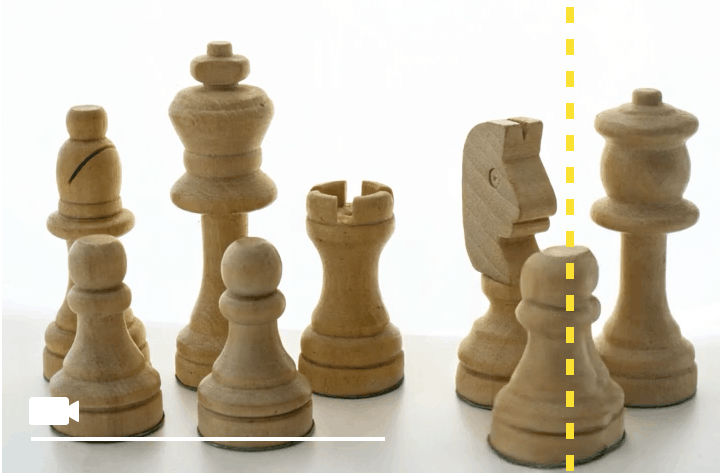}
        \includegraphics[width=0.22\linewidth]{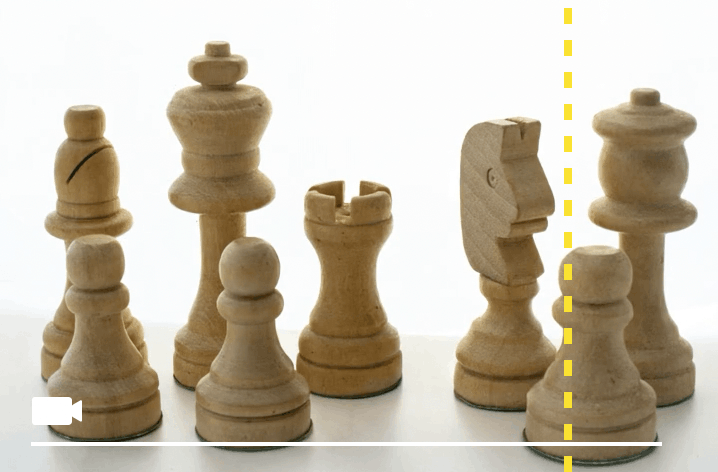}
        \end{minipage}
    }
    \caption{\textbf{Failure Cases.} We noticed that KineMask fails to generate collisions with objects that do not have a considerable height and in ambiguous setups.}
    \label{fig:Failure}
    \vspace{-10pt}
\end{figure*}

Figure~\ref{fig:Failure} shows some failure cases of \methodname. In the first and second case, we have found that objects that do not have a considerable height are susceptible to ignore others thus a collision is not created. The third and fourth cases, show complex scenarios with many similar objects. This sometimes creates ambiguities, by resulting in object duplication or disappearance. We speculate here that the text prompt may also encourage ambiguity, due to the presence of multiple elements that can be associated to the same textual identifier. \looseness=-1


\end{document}